\documentclass[10pt,twocolumn,letterpaper]{article}

\usepackage[pagenumbers]{cvpr} % To force page numbers, e.g. for an arXiv version
\usepackage[dvipsnames]{xcolor}

\usepackage{graphicx}
\usepackage{amsmath}
\usepackage{amssymb}
\usepackage{booktabs}
\usepackage{enumitem}
\usepackage[misc]{ifsym}
\usepackage{multirow}

\definecolor{cvprblue}{rgb}{0.21,0.49,0.74}
\usepackage[pagebackref,breaklinks,colorlinks,citecolor=cvprblue]{hyperref}
\usepackage[toc,page]{appendix}
\graphicspath{ {./data/} }

\usepackage[capitalize]{cleveref}
\crefname{section}{Sec.}{Secs.}
\Crefname{section}{Section}{Sections}
\Crefname{table}{Table}{Tables}
\crefname{table}{Tab.}{Tabs.}

\def\paperID{*****} % *** Enter the Paper ID here
\def\confName{CVPR}
\def\confYear{2024}

\begin{document}

\title{I2VEdit: First-Frame-Guided Video Editing via \\Image-to-Video Diffusion Models}
%%%%%%%%% AUTHORS - PLEASE UPDATE
\author{
Wenqi Ouyang$^{1}$,
Yi Dong$^{3}$,
Lei Yang$^{2}$,
Jianlou Si$^{2}$,
Xingang Pan$^{1}$
\\
$^{1}$S-Lab, Nanyang Technological University,\\
$^{2}$SenseTime Research and Shanghai AI Laboratory,\\
$^{3}$Nanyang Technological University
\\
{\tt\small \{wenqi.oywq, ydong004, xingang.pan\}@ntu.edu.sg,}\\
{\tt\small \{yanglei, sijianlou\}@sensetime.com}
}

\twocolumn[{%
\renewcommand\twocolumn[1][]{#1}%
\maketitle
\vspace{-2.3em}
\begin{center}
\centering
\includegraphics[width=\linewidth]{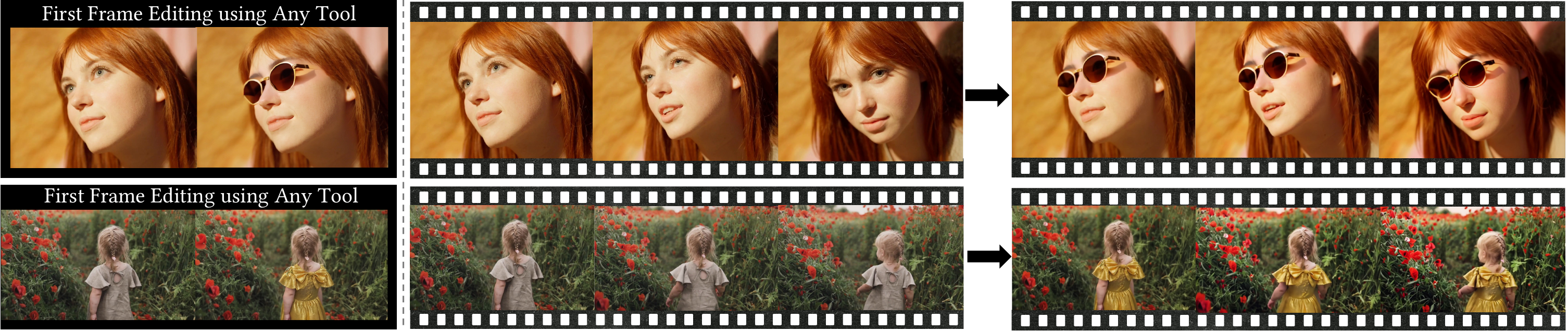}%\vspace{-0.4em}
\captionof{figure}{Our video editing pipeline. Given the first frame edited by users using an image editing tool (\eg, EditAnything~\cite{gao2023editanything}), our model generates videos consistent with first frames, while preserving appearances and motion adaptively with source videos.}
\label{fig:1_1}
\end{center}%
}]

\maketitle

\begin{abstract}
  %Video editing aims to tailor video appearances to specific user requirements. Previous methods, leveraging text-guided image and video models, have been limited to a narrow scope of edits, and fall short in tasks requiring precise local edits and concept customization. 
  %%We introduce a novel framework that leverages a pre-trained image-to-video model for editing videos based on arbitrarily user-edited first frames, adaptively maintaining the source video's visual and motion integrity relative to the editing extent.
  %In light of the gap between image editing and video editing, we introduce a novel framework that, based on a user-edited first frame via any image editing tools, propagates such editing to other frames via a pre-trained image-to-video model and adaptively maintains the source video's visual and motion integrity depending on the editing extent.
  The remarkable generative capabilities of diffusion models have motivated extensive research in both image and video editing. Compared to video editing which faces additional challenges in the time dimension, image editing has witnessed the development of more diverse, high-quality approaches and more capable software like Photoshop. In light of this gap, we introduce a novel and generic solution that extends the applicability of image editing tools to videos by propagating edits from a single frame to the entire video using a pre-trained image-to-video model. Our method, dubbed I2VEdit, adaptively preserves the visual and motion integrity of the source video depending on the extent of the edits, effectively handling global edits, local edits, and moderate shape changes, which existing methods cannot fully achieve.
  %Our method, dubbed I2VEdit, features two main processes: 
  At the core of our method are two main processes: Coarse Motion Extraction to align basic motion patterns with the original video, and Appearance Refinement for precise adjustments using fine-grained attention matching. We also incorporate a skip-interval strategy to mitigate quality degradation from auto-regressive generation across multiple video clips. %allowing for the generation of videos up to 117 frames in length. % 117 does not seems to be very long, no need to highlight this in abstract.
  Experimental results demonstrate our framework's superior performance in fine-grained video editing, proving its capability to produce high-quality, temporally consistent outputs. Our website is at \url{https://i2vedit.github.io/}.
\end{abstract}

\section{Introduction}
\label{sec:1}

In recent years, video has emerged as an increasingly popular and important medium for conveying information.
As the demand for high-quality video content grows, so does the need for sophisticated video editing tools. 
Recent advancements in image and video diffusion models~\cite{rombach2021highresolution,guo2023animatediff,blattmann2023stable,wang2023modelscope} have shown tremendous potential for automatic video editing, promising to significantly reduce the manual labor traditionally required in the field.
An ideal video editing tool should be capable of performing a wide range of edits including global edits, such as style transfer, and local edits, such as replacing or revising specific objects without affecting other contents, to cater to the diverse needs of media content creators. %designers.
% Do we want to mention somewhere that we focus on spatial/appearance editing?

While significant progress has been made in video editing using diffusion models, existing methods are often restricted to a limited subset of editing tasks.
For example, a series of works extends pre-trained text-to-image models to achieve video editing using strategies to keep temporal consistency and preserve spatial layouts, including attention manipulation~\cite{liu2023videop2p,qi2023fatezero,text2video-zero,vid2vid-zero,zhang2023controlvideo}, guidance by optical flows or depth maps~\cite{yang2023rerender,chen2023eve,jeong2023ground,cong2023flatten}, and one-shot tuning~\cite{wu2023tune,zhao2023controlvideo}. 
Despite demonstrating a certain degree of editing capability, these methods mainly focus on global style transfer with little structural change or fail to achieve fine-grained local editing without affecting irrelevant areas.
Another line of research in motion customization can synthesize videos with motions close to the source video based on text-guided video diffusion models~\cite{jeong2023vmc,zhao2023motiondirector,yatim2023spacetime}, but with limited capability to keep spatial appearance consistent with source videos.

Unlike video editing tasks that face the additional challenge of handling temporal consistency and cross-frame spatial correlations, image editing tasks have fewer constraints and have witnessed much more rapid developments.
Powerful image editing tools across a wide range of varieties have been developed including general editing methods~\cite{couairon2022diffedit,hertz2022prompt,cao_2023_masactrl,brooks2022instructpix2pix}, concept customization~\cite{ruiz2023dreambooth,dong2022dreamartist,gu2023mixofshow,ye2023ip-adapter,xiao2023fastcomposer}, fine-grained local editing guided by semantic masks~\cite{gao2023editanything,chen2023anydoor}, and established commercial software like Photoshop~\cite{adobephotoshop}.
The substantial gap between image and video editing motivates us to explore a powerful yet much less explored way of video editing, which is to \textit{edit the first frame using any powerful image editing tools and then propagate such edits to other frames} via a pre-trained image-to-video model~\cite{blattmann2023stable}, as shown in \cref{fig:1_1}.
This strategy divides the problem of content editing and preservation of motion and temporal consistency, enabling us to leverage any off-the-shelf powerful image editing tools for video editing.

To this end, we present I2VEdit, a video editing approach guided by first frame editing that adaptively preserves the spatial appearance and motion trajectories of source videos depending on the extent of edits. %regardless of how the first frame is edited.
%To achieve this, our framework consists of three key components.
This task involves several challenges, addressed by four key components in our framework.
\textbf{a)} To preserve the motion of the source video in the output, we start by training a motion LoRA that captures the coarse motion in the source video.
\textbf{b)} Our approach further refines the appearance and motion by fine-grained attention matching, which adaptively adjusts its strength to handle different levels of structural changes.
%c) As b) is based on DDIM inversion~\misscite, we propose a smooth area random perturbation (SARP) technique that significantly improves DDIM inversion and extracts more meaningful latents and attentions for source videos with large smooth regions.
\textbf{c)} Observing that deterministic EDM~\cite{Karras2022edm} and DDIM~\cite{song2020denoising} inversion sampling, which lays the foundation for b), often fails for source videos with large smooth regions, we propose a smooth area random perturbation (SARP) technique that significantly improves inversion sampling and extracts much more meaningful latents and attentions.
%\textbf{d)} We also devise a skip-interval approach to minimize the quality decline resulting from the auto-regressive strategy for long video generation, generating video with a max length of 118 frames on UBC Fashion dataset~\cite{zablotskaia2019dwnet}.
\textbf{d)} We also devise a skip-interval approach that enables us to apply the auto-regressive strategy for long video editing with significantly less quality degradation.

We extensively evaluate I2VEdit and demonstrate that it effectively extends existing image editing methods to the video domain.
Thanks to the rich and powerful image editing tools, I2VEdit offers greater flexibility compared to other video editing methods, especially in terms of local edits, as shown in \cref{fig:1_1}.
The visual editing quality is also enhanced due to the superior base image editing method.
When compared to another image-guided video editing method, Ebsynth~\cite{jamrivska2019stylizing}, our method produces starkly more realistic results with much fewer artifacts.
To summarize, our contributions are as follows:
\begin{itemize}[itemsep=0pt,topsep=0pt,parsep=0pt]
\item We propose a novel framework, I2VEdit, to achieve fine-grained video editing based on the pre-trained image-to-video model. Given a first frame edited arbitrarily by users using any powerful image editing tools, our framework generates video consistent with the first frame, adaptively preserving the visual appearance and motion trajectories of source videos based on the extent of edits. 
\item We match the coarse motion of output with the source video by training skip-interval motion LoRAs, while also reducing the quality decline resulting from the auto-regressive generation strategy. %We generate edited videos with a max length of 118 frames on UBC Fashion Dataset~\cite{zablotskaia2019dwnet}.  % xingang: 118 frames is not much, so I remove this sentence.
\item We design fine-grained attention-matching algorithms to adaptively match appearance and motion with source videos via spatial attention difference map calculation and multi-stage temporal attention injection. We also propose smooth area random perturbation for deterministic EDM and DDIM inversion sampling to improve the editing quality of videos with large constant pixel areas.
\end{itemize}

\section{Related Work}
\label{sec:2}
%\subsection{Image Editing with Diffusion Models}
%\label{sec:2-1}
\noindent {\bf Image Editing with Diffusion Models.}
Image editing involves generating images based on reference images and textual prompts, ensuring alignment with both references and textual commands. Many attempts have been made to achieve this task using the pre-trained text-to-image model, \eg, Stable Diffusion~\cite{rombach2021highresolution}. These approaches can be broadly divided into three categories: zero-shot image editing ~\cite{couairon2022diffedit,hertz2022prompt,cao_2023_masactrl,chen2023anydoor}, methods with one-shot tuning~\cite{ruiz2023dreambooth,dong2022dreamartist,gu2023mixofshow} and large-data-driven methods~\cite{gao2023editanything,brooks2022instructpix2pix,ye2023ip-adapter,xu2023prompt}. Prompt-to-Prompt~\cite{hertz2022prompt} and MasaCtrl~\cite{cao_2023_masactrl} modify attention maps according to the textual tokens to achieve zero-shot image editing. Instruct-Pix2Pix~\cite{brooks2022instructpix2pix} generates training data using Prompt-to-Prompt, training an editing model by utilizing referenced images and text prompts as model input. IP-Adapter~\cite{ye2023ip-adapter} trains an image encoder to preserve features from the original image, generating editing results with similar appearances. EditAnything~\cite{gao2023editanything} further improves editing accuracy with semantic and user-drawn masks as control conditions, preserving the original appearance while generating high-quality editing results. There are also well-developed editing approaches for specific applications, such as virtual try-on~\cite{lee2022hrviton,idm-vton}, and concept customization for humans~\cite{xiao2023fastcomposer}. We employ EditAnything~\cite{gao2023editanything}, AnyDoor~\cite{chen2023anydoor}, Instruct-Pix2Pix~\cite{brooks2022instructpix2pix}, InstantStyle~\cite{wang2024instantstyle}, and IDM-VTON~\cite{idm-vton} as the primary first-frame editing tools in most of our experiments.

%\subsection{Text-Guided Video Editing and Motion Customization}
%\label{sec:2-2}
\noindent {\bf Text-Guided Video Editing and Motion Customization.}
Text-guided video editing aims to adjust the visual appearance of videos based on textual prompts while preserving the original video's characteristics. Prior approaches have utilized pre-trained text-to-image models for achieving zero-shot video editing. These methods can be categorized into two groups based on their approach to maintaining temporal consistency: methods modifying attention mechanisms for cross-frame correlations~\cite{liu2023videop2p,qi2023fatezero,text2video-zero,vid2vid-zero,zhang2023controlvideo}, and those incorporating constraints from optical flows or depth maps~\cite{yang2023rerender,chen2023eve,jeong2023ground,cong2023flatten}. FateZero~\cite{qi2023fatezero} achieves appearance and shape editing through attention fusion guided by textual tokens but lacks fine-grained editing control. Rerender-A-Video~\cite{yang2023rerender} produces high-quality frames but is limited to global style transfer with minimal structural variation due to optical flow alignment. Other methods introduce temporal layers into text-to-image models and fine-tune them on individual videos to learn temporal correlations, such as Tune-A-Video~\cite{wu2023tune} and ControlVideo~\cite{zhao2023controlvideo}, yet they may suffer from reduced editing quality due to overfitting to the original video. 

Text-guided motion customization aims to generate videos that not only mirror the motion of original videos but also align seamlessly with text prompts, such as VMC~\cite{jeong2023vmc}, MotionDirector~\cite{zhao2023motiondirector} and Space-Time Features~\cite{yatim2023spacetime}. These methods generate videos with motion trajectories roughly matched with the original videos at a coarse level, lacking precise editing capability for visual appearances in the generated results. 

%\subsection{Image-to-Video Generation and Editing}
%\label{sec:2-3}
\noindent {\bf Image-to-Video Generation and Editing.}
Previous methods design hand-crafted algorithms to perform example-based video stylizing, such as Ebsynth~\cite{jamrivska2019stylizing}. However, it suffers from limited generation quality in shape variation and structural changes. Recently, diffusion models have been used to solve image-to-video generation problems, such as Stable Video Diffusion~\cite{blattmann2023stable}, Gen-2~\cite{gen-2}, I2Vgen-XL~\cite{2023i2vgenxl}, PikaLabs~\cite{pika}, SparseCtrl~\cite{guo2023sparsectrl} and SORA~\cite{sora2024}. DragNUWA~\cite{yin2023dragnuwa}, MoVideo~\cite{liang2023MoVideo} further control the generation of videos with explicit optical flows or trajectories. Other concurrent works, such as MotionI2V~\cite{shi2024motion}, MoCA~\cite{yan2023motion} and MagicProp~\cite{yan2023magicprop}, achieve video editing with similar strategies to ours, \eg, using an edited keyframe to guide the editing process. However, they rely on optical flows or depth maps of source videos to control the image-to-video generation process. CoDef~\cite{codef} generates edited videos by propagating edits from the first frame using deformation fields extracted from the source videos. However, it faces challenges when dealing with editing cases involving local objects and structural changes. VideoSwap~\cite{gu2023videoswap} utilizes sparse key points of source videos to control motion trajectories of generated foreground subjects. Its editing capabilities may be constrained by explicit guidance, limiting fine-grained control over shapes and appearances. AnyV2V~\cite{ku2024anyv2v} utilizes image-to-video models to edit videos given the edited first frame by a training-free strategy. However, it may generate results with temporal inconsistency and structural changes. In contrast, our method enables users to perform any desired edits on the first frame, generating videos aligned with it while preserving the appearances and motion of source videos adaptively based on the extent of edits.

\section{Preliminaries}
\label{sec:3}

\noindent{\bf Image-to-Video Diffusion Model.}
Image-to-video diffusion model, \eg, Stable Video Diffusion~\cite{blattmann2023stable}, designs a 3D U-Net with temporal convolution and attention structures to generate videos from Gaussian noises, guided by a conditional image as the first frame of the output video. The conditional image is encoded into CLIP~\cite{radford2021learning} image embedding for cross-attention. Additionally, a noise-augmented version of the conditional image is concatenated channel-wise with the input of the 3D U-Net. The 3D U-Net is optimized by the loss with EDM noise schedule~\cite{Karras2022edm}:
 \begin{equation}
\begin{aligned}
z_{t,\sigma} = [z_{t}; c_{\sigma}], c_{\sigma} = &\tau(c+\sigma) \\
L_{svd} = \mathbb{E}_{z_{0},c,\epsilon\sim\mathcal{N}(0,I),t\sim\mathcal{U}(0,T)}&[\lVert z_{0}-z_{\theta}(z_{t,\sigma},t,c)\rVert^{2}_{2}]
 \end{aligned}
  \label{eq:3.1}
\end{equation}
where $c$ is the conditional image, $\sigma$ represents the Gaussian noise, %\lei{and $\tau(\cdot)$ denotes xxx}.
and $c_{\sigma}$ represents the noise-augmented conditional image encoded by VAE encoder $\tau(\cdot)$. 

\noindent{\bf Low-Rank Adaptation.} Low-rank adaptation (LoRA)~\cite{hu2021lora} presents a novel framework designed to efficiently fine-tune large language models with a small subset of the parameters for task-specific adaptation. It can be applied for video models to achieve motion customization and control, such as MotionDirector~\cite{zhao2023motiondirector} and DragNUWA for SVD~\cite{yin2023dragnuwa}. For a pre-trained weight matrix $W_{0} \in \mathbb{R}^{d \times k}$, LoRA constrains its update by a low-rank decomposition:
 \begin{equation}
\begin{aligned}
W_{0} + {\rm \Delta} W=W_{0} + BA
 \end{aligned}
  \label{eq:3.2}
\end{equation}
where $B \in \mathbb{R}^{d\times r}, A \in \mathbb{R}^{r \times k} $. Rank $r$ is much smaller than $d$ and $k$.
\section{Approach}
\label{sec:4}
\begin{figure*}[tb]
    \centering
    \includegraphics[width=\textwidth]{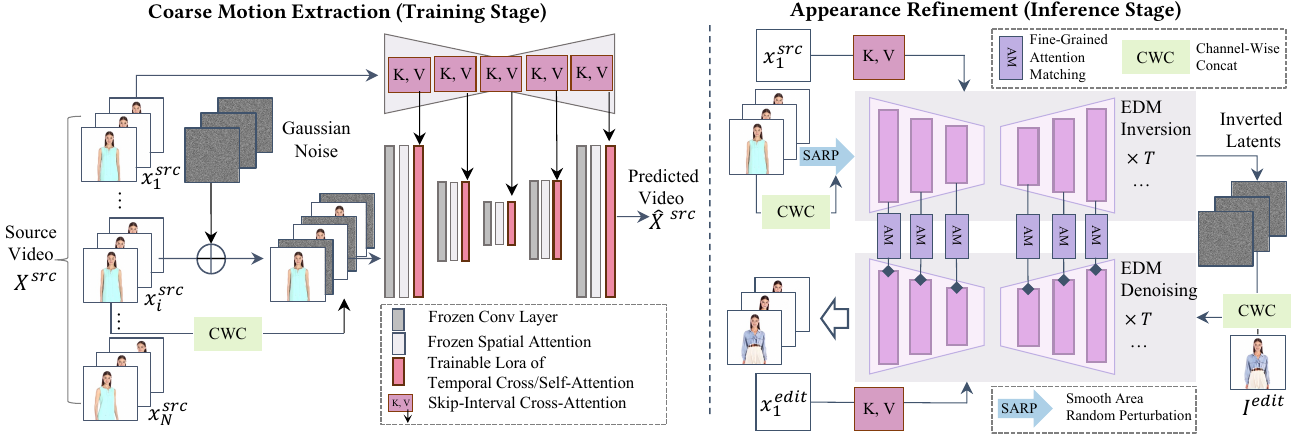}
    \vspace{-0.3cm}
    \caption{Our framework comprises two pipelines: Coarse Motion Extraction Pipeline (Training Stage) and Appearance Refinement Pipeline (Inference Stage). Coarse Motion Extraction Pipeline extracts coarse motion via learning skip-interval motion LoRAs for each clip. In the inference stage, Appearance Refinement Pipeline further refines the motion and appearance consistency through fine-grained attention matching between attentions during EDM~\cite{Karras2022edm} inversion and denoising.}
    %\vspace{-0.1cm}
    \label{fig:4_1}
\end{figure*}
%Skip-Interval Motion Lora, Smooth Area Random Perturbation, and Multi-Scale Attention Matching. For a given long video, it is initially segmented into clips of lengths suitable for video model. In the training stage, we implement a skip-interval approach for each clip to efficiently train the motion lora, mitigating the performance decline associated with auto-regressive strategy and capturing motion details. In the inference stage, we employ DDIM inversion alongside smooth area random perturbation on each clip, promoting the generation of inverted latents that adhere more closely to a Gaussian distribution. Subsequently, Multi-Scale Attention Matching is applied within the DDIM denoising step to ensure fidelity in motion and appearance to the original clip. This section will detail the functionality and significance of each module.
%Our video editing approach adopts a first-frame-guided strategy, generating an edited video $\hat{X}^{edit}$ from a source video $X^{src}$ and an edited first frame $I^{edit}$, preserving motion and appearance adaptively. 
Given a source video $X^{src}$ and an edited first frame $I^{edit}$ obtained via an image editing tool, our method generates an edited video $\hat{X}^{edit}$ that is consistent with $I^{edit}$.
The type of editing can be either local or global and can involve both appearance and moderate shape changes.
The motion of the edited video should align with the source video, while editing the motion itself falls outside the scope of this work.
We utilize a pre-trained image-to-video model, \eg, Stable Video Diffusion~\cite{blattmann2023stable}, as the base model. The whole framework comprises two pipelines: Coarse Motion Extraction and Appearance Refinement, as shown in \cref{fig:4_1}. The source video $X^{src}$ is initially segmented into $N$ clips $\{x^{src}_{1},..,x^{src}_{N}\}$, each of a length suitable for the image-to-video model. Coarse Motion Extraction Pipeline extracts coarse motion from the source video by learning motion LoRAs for each clip, along with skip-interval cross-attention to mitigate performance decline associated with the first-frame-conditioning auto-regressive strategy of the image-to-video model. Appearance Refinement pipeline further enhances motion and appearance consistency between each clip pair $x^{src}_{i}$ and $x^{edit}_{i}$. The following sections detail each pipeline's functionality and significance.

\subsection{Coarse Motion Extraction}
\label{sec:4.1}
\noindent {\bf Motion LoRA.}
To capture the coarse motion of $x^{src}_{i}$, we fine-tune the video model by adding LoRAs to the temporal attention layers, similar to MotionDirector~\cite{zhao2023motiondirector}. However, we refrain from using spatial LoRAs and the appearance-debiased temporal loss, as they destabilize the training process and often yield unsatisfactory outcomes with image-to-video models like Stable Video Diffusion~\cite{blattmann2023stable}. Further details are provided in \cref{app:motionlora}. Notably, the image-to-video model demonstrates a sufficient capability to align the appearance of the generated video with the conditional image without the need for additional spatial LoRAs.

\noindent {\bf Skip-Interval Cross-Attention.}
For the first-frame-conditioning image-to-video model, an auto-regressive strategy can be used to generate a long video, using the last frame of the previous clip as the conditional image for the current clip. However, this will result in a performance decline due to the information loss and quality gap between the last generated frame of each clip and the initial keyframe. In order to reduce the performance decline, in the training stage, we perform EDM~\cite{Karras2022edm} inversion sampling on $x^{src}_{1}$ (\ie, reverse process of EDM denoising), saving key and value matrices of temporal self-attention for each step. When training motion LoRAs for other clips $\{x^{src}_{i}\}^{N}_{i=2}$,  these matrices are concatenated with key and value matrices of current temporal self-attention for each step, enabling skip-interval cross-attention with $x^{src}_{1}$. Since key and value matrices contain the appearance features, this strategy can help preserve the original appearance of the edited image. The output of temporal self-attention $\mathbf{Z}^{s}$ with skip-interval cross-attention is represented as follows:
\begin{equation}
\begin{aligned}
\mathbf{K}^{s} =[\mathbf{K}';\mathbf{K}]&, \mathbf{V}^{s}=[\mathbf{V}';\mathbf{V}] \\
\mathbf{Z}^{s} ={\rm Attention}(\mathbf{Q',K}^{s},\mathbf{V}&^{s})={\rm softmax}(\frac{\mathbf{Q}'(\mathbf{K}^{s})^{T}}{\sqrt{d}})\mathbf{V}^{s}
  \end{aligned}
  \label{eq:4.1_1}
\end{equation}
where $\mathbf{Q',K',V'}$ are the query, key, and value matrices of temporal self-attention for current clip $x^{src}_{i}$, $\mathbf{K,V}$ are the key and value matrices for $x^{src}_{1}$. In the inference stage for editing, $\mathbf{K,V}$ are generated during the denoising process of $x^{edit}_{1}$ to perform skip-interval cross-attention with $\{x^{edit}_{i}\}^{N}_{i=2}$.

\vspace{0.5em}
\noindent {\bf Training Strategy.}
We implement a similar training strategy to the image-to-video model, \eg, Stable Video Diffusion~\cite{blattmann2023stable} with EDM noise schedule~\cite{Karras2022edm} and svd-temporal-controlnet~\cite{svdcontrolnet}. 
%In order to reduce quality decline when generating long videos using an auto-regressive strategy, 
In order to accelerate training process, we utilize caching latents by adding noise $\sigma$ to the conditional latents $\tau(c)$ obtained by encoding the conditional image $c$. The conditional image latents $c_{\sigma}$ are concatenated channel-wise with noise-augmented input video latents $z_{t}$, resulting in $z_{t,\sigma}$ as input to the video model. The loss function used to train LoRA is represented as follows:
%\lei{Please check Eq.~\label{eq:3.1}, where it is $c_{\sigma} = \tau(c + \sigma)$}
\begin{equation}
\begin{aligned}
L_{motion} = \mathbb{E}_{z_{0},c,\epsilon\sim\mathcal{N}(0,I),t\sim\mathcal{U}(0,T)}&[\lVert z_{0}-z_{\theta}(z_{t,\sigma},t,c)\rVert^{2}_{2}], \\
\text{where } z_{t,\sigma} = [z_{t}; c_{\sigma}], c_{\sigma} = &\tau(c) + \sigma
 \end{aligned}
  \label{eq:4.1_2}
\end{equation}
\subsection{Appearance Refinement}
\label{sec:4.2}
To enhance the alignment of motion and appearance with the source video, we begin by performing EDM inversion of $x^{src}_{i}$, storing spatial and temporal self-attentions. EDM denoising is then conducted using the inverted latents which contain more specific motion information, along with the edited keyframe as a condition to obtain $x^{edit}_{i}$. During the denoising process, fine-grained attention matching rectifies spatial and temporal self-attentions based on pre-saved attentions obtained from inversion, ensuring adaptive preservation of motion and appearance consistency with $x^{src}_{i}$. These modules are detailed in this section.

\noindent {\bf Smooth Area Random Perturbation (SARP).}
 To ensure the appearance of the edited clip is consistent with the edited keyframe, the inverted latents should contain less appearance information of the source clip and more closely follow Gaussian distribution to ensure no violation of the denoising process of the image-to-video model. We find that adding small perturbations in the pixel domain to the smooth area of the source clip, especially areas with constant pixel values, \eg, constant white background, would generate more Gaussian-distributed inverted latents, remarkably improving the editing quality. 
 We conjecture this is because the U-Net never sees an image with noise-free smooth areas during training, leading to a domain gap during EDM inversion. And SARP significantly addresses this domain gap issue.
 Specifically, we first detect the smooth area of the source clip using Sobel gradient thresholding~\cite{kanopoulos1988design} to obtain the mask for smooth area $M_{sarp}$, then add small noise on the source clip $x^{src}_{i}$:
\begin{equation}
\begin{aligned}
x^{src}_{sarp} = (x^{src}_{i} + \alpha \cdot \epsilon)& \odot M_{sarp} + x^{src}_{i} \odot (1-M_{sarp}), \\
\epsilon &\sim \mathcal{N}(0,1)
  \end{aligned}
  \label{eq:4.2_1}
\end{equation}
where $\alpha$ is the noise scale, which is a relatively small value compared with pixel values of source clip.

\begin{figure}[tb]
    \centering
    \includegraphics[width=0.42\textwidth]{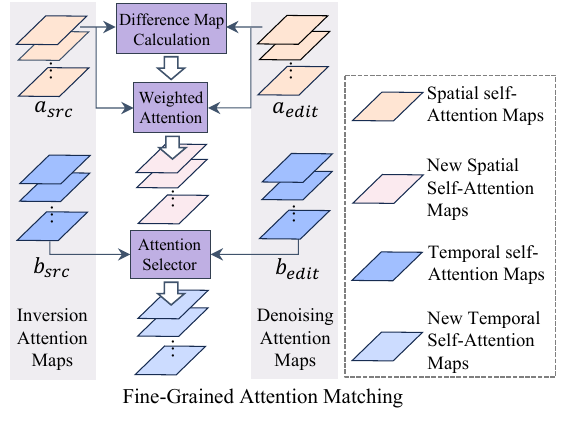}
    \vspace{-0.2cm}
    \caption{Fine-Grained Attention Matching.}
    \vspace{-0.2cm}
    \label{fig:4_2}
\end{figure}

\noindent {\bf Fine-Grained Attention Matching.}
We implement an attention matching strategy to refine appearances of edited videos, as shown in \cref{fig:4_2}. During the inversion of the source clip, we store the inverted latent $z_T$ and intermediate self-attention maps as:
\begin{equation}
\begin{aligned}
z_T, \{a^{src}_{t}\}^{T}_{t=0}, \{b^{src}_{t}\}^{T}_{t=0} = \text{EDM-INV}(x^{src}_{sarp})
 \end{aligned}
 \label{eq:4.2_2}
\end{equation}
%During the inversion of the source clip $x^{src}_{sarp}$, we store spatial and temporal self-attention maps for each denoising time step $t$ as $\{a^{src}_{t}\}^{T}_{t=0}$ and $\{b^{src}_{t}\}^{T}_{t=0}$ respectively.
where EDM-INV stands for the EDM inversion process, $a^{src}_{t}$ and $b^{src}_{t}$ are spatial and temporal self-attention maps for time step $t$ of source video clip, respectively.

The edited video $X^{edit}$ is generated by performing EDM denoising process based on the inverted latent $z_T$ conditioned on the edited image $I^{edit}$, together with the attention matching mechanism as below.
%During EDM denoising process, 
For each time step $t$, we calculate the difference map between $a^{src}_{t}$ and spatial self attentions $a^{edit}_{t}$ for edited video clip. We further aggregate the difference map in the channel dimension and normalize it by a factor of $2$ to ensure its range is in $[0,1]$:
\begin{equation}
\begin{aligned}
a^{diff}_{t} &= \left | a^{edit}_{t} - a^{src}_{t} \right | \\
{\hat a}^{diff}_{t} &= \sum_{c} a^{diff}_{t,c} / 2
  \end{aligned}
  \label{eq:4.2_3}
\end{equation}
Since spatial self-attention maps contain structural information of frames, ${\hat a}^{diff}_{t}$ indicates structural differences between source frames and edited frames. For local editing tasks, a higher value means the generation of new edited objects, while a lower value indicates the unedited area that should preserve consistency with source frames. As for global editing tasks, \eg, style transfer, a lower value indicates little structural changes despite global style variation of appearance. We use ${\hat a}^{diff}_{t}$ to generate weighted attention $a^{w}_{t}$ for edited frames:
 \begin{equation}
\begin{aligned}
&M^{diff}_{t} = \left \{
\begin{aligned}
1, {\hat a}^{diff}_{t} > {\rm thr} \\
{\hat a}^{diff}_{t}, {\hat a}^{diff}_{t} \leq {\rm thr}
 \end{aligned}
 \right. \\
a^{w}_{t} &= a^{edit}_{t} \odot M^{diff}_{t} + (1 - M^{diff}_{t}) \odot a^{src}_{t}
  \end{aligned}
  \label{eq:4.2_4}
\end{equation}
where {\rm thr} represents the value for thresholding. Original attentions $a^{edit}_{t}$ for edited frames are replaced by $a^{w}_{t}$, matching motions and appearances with source frames.
For temporal self-attention,  we use the attention selector to modify attention maps $b^{edit}_{t}$ for edited frames. Specifically, we divide the denoising process by time steps into three stages. In the first stage $t \in [0.0, \beta_{1}\times T)$, $b^{edit}_{t}$ is directly replaced by $b^{src}_{t}$. In the second stage $t \in [\beta_{1}\times T, \beta_{2}\times T)$, only the attentions with large downscaling factors are replaced. In the last stage $t \in [\beta_{2}\times T, T]$, $b^{edit}_{t}$ are keep unmodified to preserve fine-grained edited details. 

\section{Experiments}
\label{sec:5}
\begin{figure*}[htbp]
    \centering
    \includegraphics[width=1.0\textwidth]{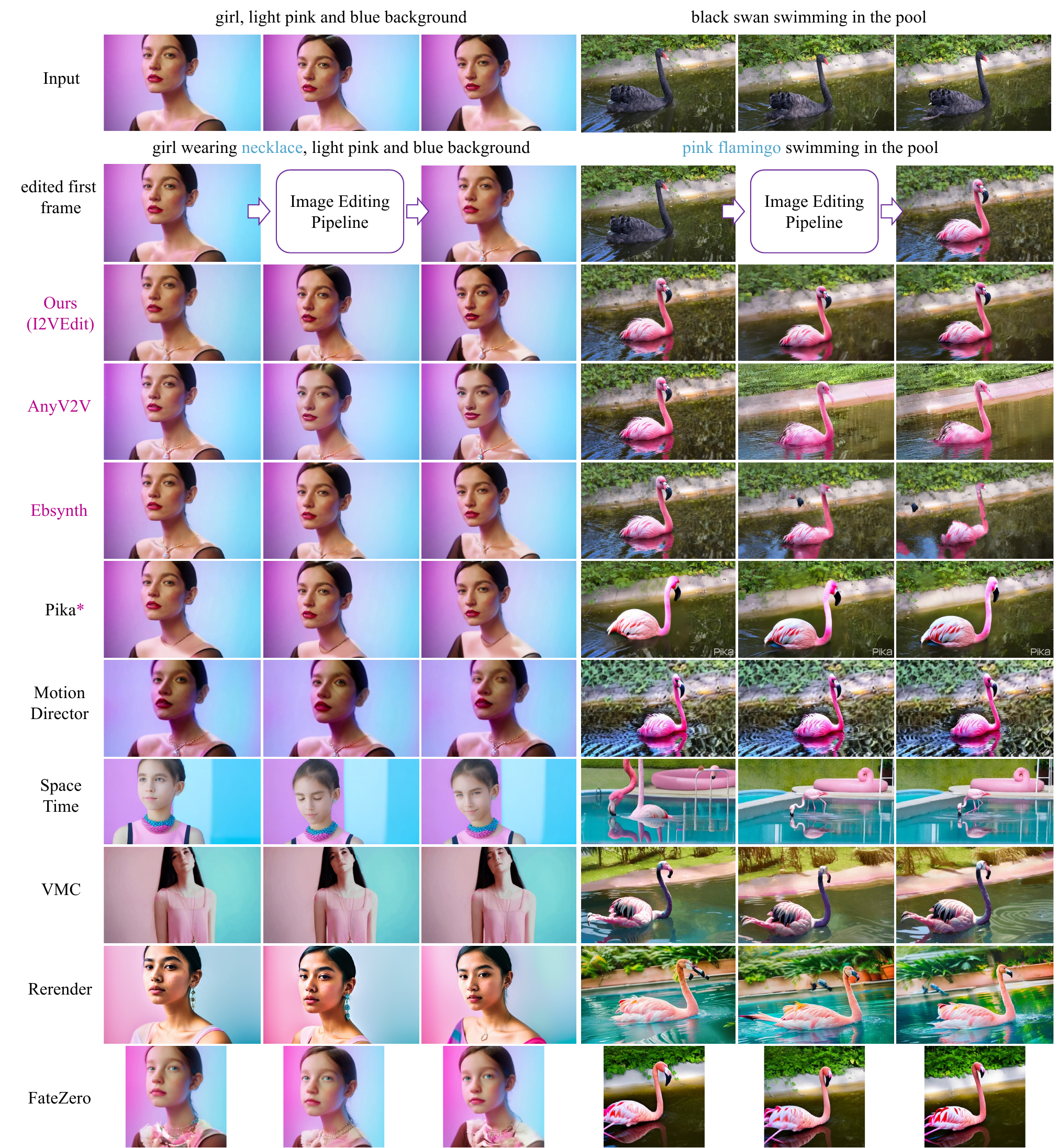}
    \vspace{-0.5cm}
    \caption{Qualitative comparison with \textcolor{purple}{image-guided video editing (colored as purple)}, text-guided video editing, and motion customization methods. We use EditAnything~\cite{gao2023editanything} to generate first-frame editing results for all image-guided video editing methods. "\textcolor{purple}{*}" means the method utilizes an additional editing mask.} %  Image-guided methods are colored as purple.
    \vspace{-0.3cm}
    \label{fig:5_1}
\end{figure*}
\begin{table*}[htbp]
  \small
    \setlength{\tabcolsep}{1.5pt} % Reduce the space between columns
\caption{Quantitative evaluation. These key aspects are motion preservation (MP), appearance alignment with source video in the unedited area (AA), editing quality (EQ), temporal consistency (TC), and appearance consistency with the first frame (AC).}
\vspace{-0.2cm}
\label{tab:5_1}
\centering
\begin{tabular}{ccccclcccclc}
\hline
\multicolumn{1}{l}{}                                    & \multicolumn{8}{c}{Human Evaluations}                                                                                          & \multicolumn{3}{c}{Automatic Evaluations} \\ \hline
\multirow{2}{*}{Method}                                 & \multicolumn{4}{c}{Local Editing}                         &  & \multicolumn{3}{c|}{Other Tasks}                             & Local Editing  &  & Other Tasks        \\ \cline{2-12} 
                                                        & MP$\uparrow$ & AA$\uparrow$ & EQ$\uparrow$ & TC$\uparrow$ &  & MP$\uparrow$ & AC$\uparrow$ & \multicolumn{1}{c|}{TC$\uparrow$} & TC$\uparrow$   &  & TC$\uparrow$ \\ \hline
Ebsynth                                                 &     0.11         &       0.12       &   0.06          &   0.08   &  &   0.21           &   0.17           & \multicolumn{1}{c|}{0.17}            & 2.38           &  & 2.38                  \\
AnyV2V                                                 &     0.18         &       0.18       &   0.22          &   0.19   &  &   0.11           &   0.10           & \multicolumn{1}{c|}{0.10}            & 2.38           &  & 2.40                  \\
Pika                                                    &  0.07            &     0.11         &      0.06        &      0.07        &  & -            & -            & \multicolumn{1}{c|}{-}            & \textbf{2.43}           &  & -                     \\
\begin{tabular}[c]{@{}c@{}}Ours\\ (w/o AM)\end{tabular} &    0.15          &    0.13          &  0.16            &   0.17           &  &  0.25            & 0.26             & \multicolumn{1}{c|}{0.27}             & 2.38           &  & \textbf{2.42}                  \\
Ours                                                    &    \textbf{0.49}          &    \textbf{0.47}         &    \textbf{0.49}          &  \textbf{0.49}            &  & \textbf{0.43}             & \textbf{0.47}             & \multicolumn{1}{c|}{\textbf{0.47}}             & 2.40           &  & \textbf{2.42}                  \\ \hline
\end{tabular}
\vspace{-0.2cm}
\end{table*}

\subsection{Implementation Details}
\label{sec:5-1} 
For test videos, we follow Render-A-Video~\cite{yang2023rerender} to collect videos from \url{https://www.pexels.com/}. The other videos for testing are from the DAVIS 2017 dataset~\cite{pont20172017} and the UBC Fashion dataset~\cite{zablotskaia2019dwnet}. 
We use the Stable Video Diffusion~\cite{blattmann2023stable} with a frame length of $14$ as the base image-to-video generation model. For coarse motion extraction, we set LoRAs with a rank of $32$. We train $250$ steps for each clip and select the 250th checkpoint for all results generation. For appearance refinement, we resize frames to resolution $576\times1024$ to fit for Stable Video Diffusion~\cite{blattmann2023stable}. We set the gradient threshold of smooth area detection as $0.001$, detecting areas with nearly constant pixel values. The noise scale $\alpha$ for random perturbation is set to $0.005$. We set ${\rm thr} = 0.35$ for attention matching of spatial attention maps, while for temporal self-attention, the best divide of stages slightly differs among different editing cases. For local editing tasks, \eg, objects editing, we set $\beta_{1}=0.5,\beta_{2}=0.8$. For global editing without dramatic shape change, \eg, global style transfer, the stages are set as $\beta_{1}=0.8,\beta_{2}=0.9$. Global editing involving significant shape changes, \eg, coarse motion transfer, the stages are set as $\beta_{1}=0.4,\beta_{2}=0.5$. We set the downscaling factor as $4$ for the second stage. All experiments are conducted using a single NVIDIA A100 GPU.
%We set ${\rm thr} = 0.35$ for attention matching of spatial attention maps, while for temporal self-attention, the best divide of stages slightly differs among different editing cases. For local editing tasks, \eg, objects editing, we set $\beta_{1}=0.5,\beta_{2}=0.8$. For global editing without dramatic shape change, \eg, global style transfer, the stages are set as $\beta_{1}=0.8,\beta_{2}=0.9$. Global editing involving significant shape changes, \eg, coarse motion transfer, the stages are set as $\beta_{1}=0.4,\beta_{2}=0.5$. We set the downscaling factor as $4$ for the second stage. All experiments are conducted using a single NVIDIA A100 GPU. Our source code will be released.
%\xingang{(Move most implementation details to supp.)}

\subsection{Comparison with State-of-the-Arts}
\label{sec:5-2}

\noindent {\bf Comparison with Text-Guided Video Editing.}
We offer visual comparisons of our method against text-guided video editing and motion customization methods on local editing tasks, as shown in \cref{fig:5_1}. Editing results of these text-guided models are generated according to the edited text prompts. 
%We encourage the reader to watch our supplemental video for a clearer comparison.
Results of our method are generated using the conditional first frames edited by EditAnything~\cite{gao2023editanything}. The baseline methods include FateZero~\cite{qi2023fatezero}, Rerender-A-Video~\cite{yang2023rerender}, VMC~\cite{jeong2023vmc}, Space-Time Features~\cite{yatim2023spacetime}, MotionDirector~\cite{zhao2023motiondirector} and PikaLabs~\cite{pika}. These methods generate videos with motion trends roughly consistent with the original video but fail to perform local editing with accurate motion and appearance consistency. PikaLabs~\cite{pika} utilizes an extra editing mask drawn via their online tools to better preserve appearance consistency in the unedited area. However, it tends to blur the structural details in the edited area, resulting in the generation of unnatural objects. In contrast, our method generates results with better editing quality, while also better preserving both appearance and motion consistency with source videos without extra editing masks. We offer another visual comparison with Rerender-A-Video~\cite{yang2023rerender}, VMC~\cite{jeong2023vmc} and PikaLabs~\cite{pika} for style transfer tasks, which are included in \cref{app:add_results}. 
%We use the first frames generated by these text-guided methods as initial keyframes to generate editing results,
These results demonstrate the capability of our method to handle global editing and style transfer tasks.

\noindent {\bf Comparison with Image-Guided Video Editing.}
We compare our method with the image-guided video editing method, Ebsynth~\cite{jamrivska2019stylizing} and AnyV2V~\cite{ku2024anyv2v}, as shown in \cref{fig:5_1}.
%These methods use the same first frame as our method. 
The same initial keyframe is used for these methods and ours. 
%For local editing tasks, 
Ebsynth well preserves appearance consistency in unedited areas but fails to handle objects with shape and structural changes. AnyV2V fails to preserve structural features and temporal consistency. More visual results are included in \cref{app:add_results} and our website at \url{https://i2vedit.github.io/}. 
%We use AnyDoor~\cite{chen2023anydoor} and EditAnything~\cite{gao2023editanything} to generate keyframe editing results in \cref{fig:5_3} and \cref{fig:5_4}, respectively. 
\begin{figure*}[tb]
  \centering
  \includegraphics[width=0.99\textwidth]{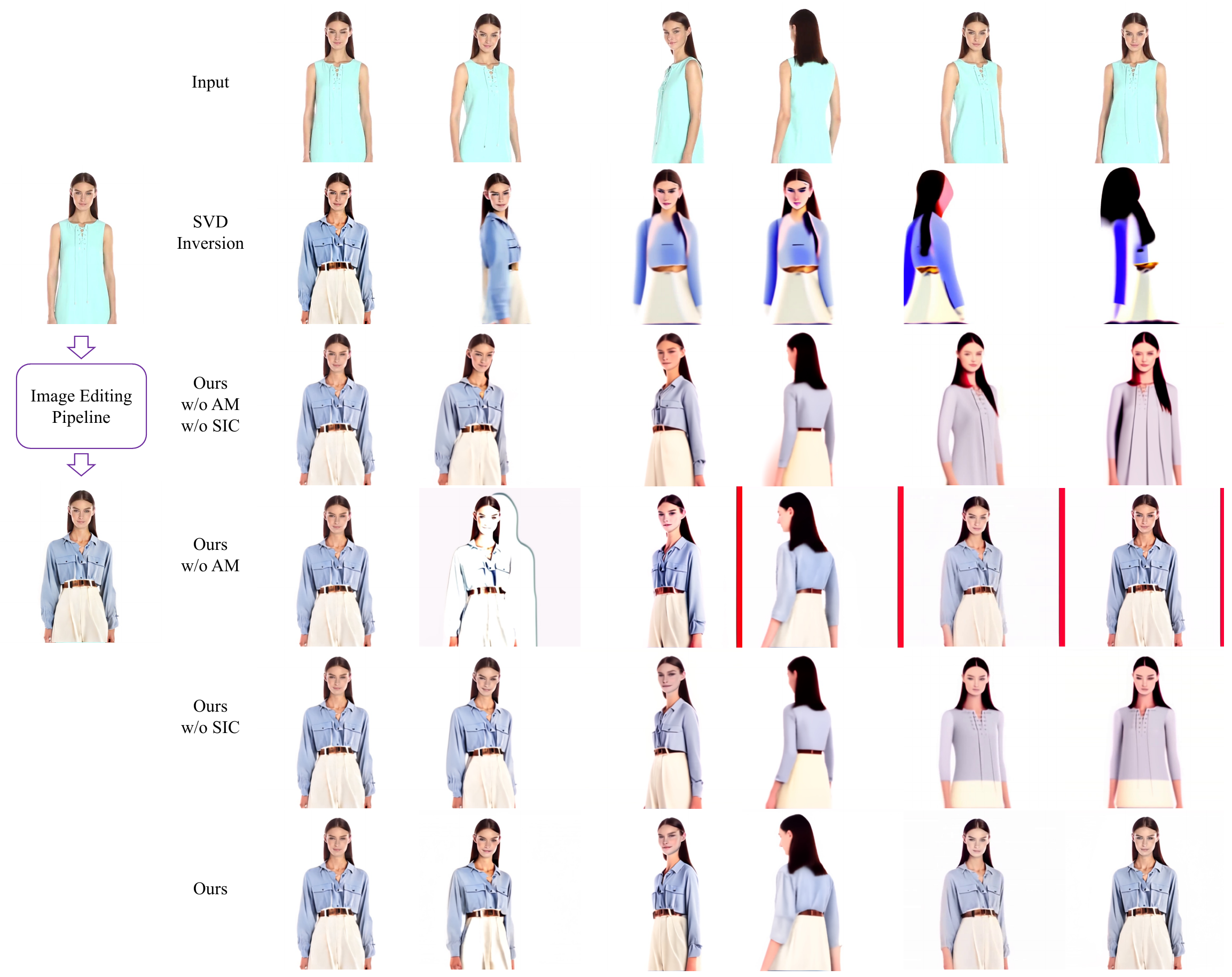}
  \vspace{-0.2cm}
  \caption{Comparison of ablation settings of our methods, using the same keyframe generated by AnyDoor~\cite{chen2023anydoor}.}
  \label{fig:5_3}
\end{figure*}
\begin{figure*}[tb]
  \centering
  \includegraphics[width=1.0\textwidth]{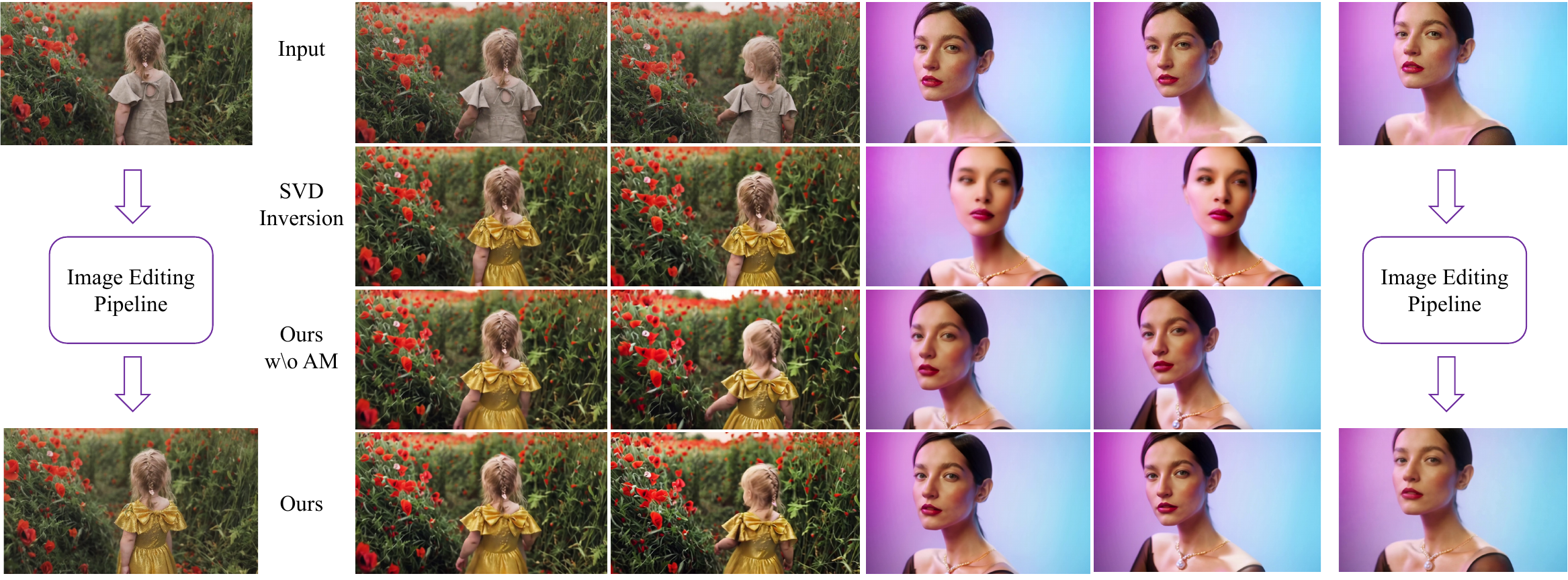}
  \vspace{-0.5cm}
  \caption{Comparison of ablation settings of our methods, using the same keyframe generated by EditAnything~\cite{gao2023editanything}.~\\~~\\~}
  \vspace{-0.2cm}
  \label{fig:5_4}
\end{figure*}

\begin{figure*}[tb]
  \centering
  \includegraphics[width=1.0\textwidth]{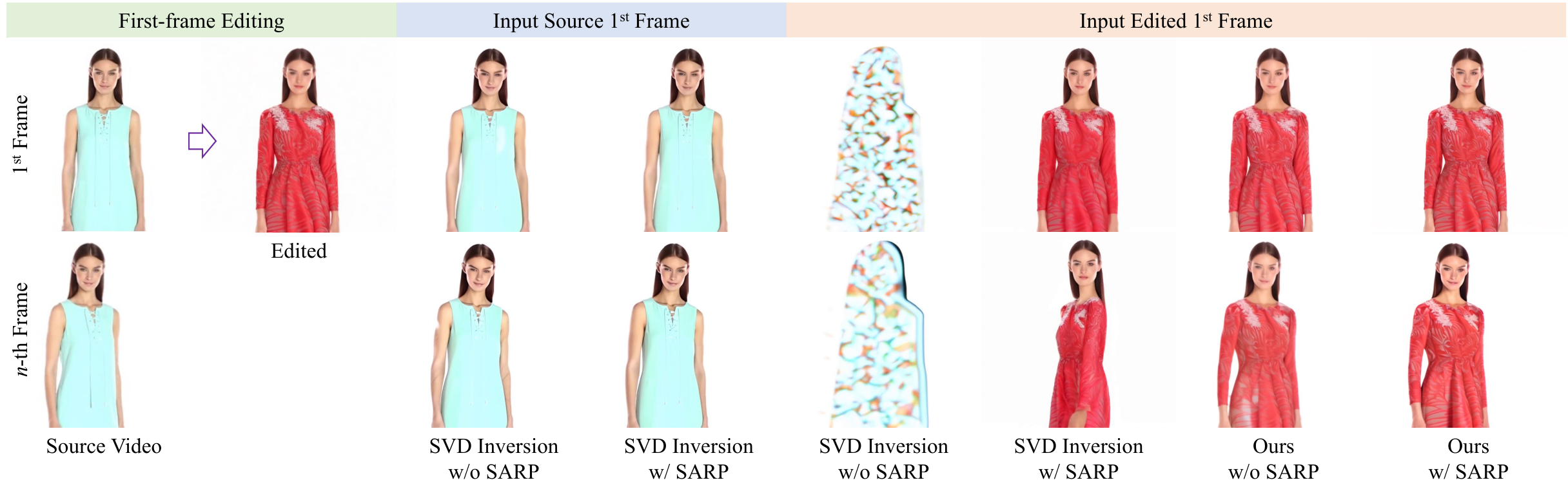}
  \vspace{-0.5cm}
  \caption{Ablation study on smooth area random perturbation for SVD, using the keyframe generated by EditAnything~\cite{gao2023editanything}.~\\~~\\~}
  \vspace{-0.2cm}
  \label{fig:5_5}
\end{figure*}

\begin{figure*}[tb]
  \centering
  \includegraphics[width=0.95\textwidth]{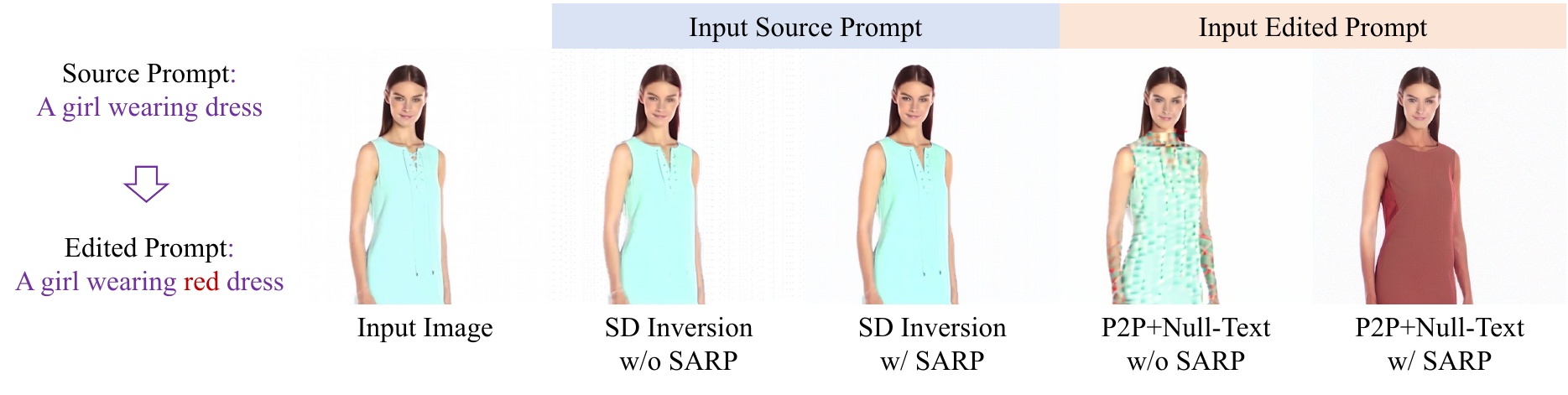}
  \vspace{-0.5cm}
  \caption{Ablation study on smooth area random perturbation for SD and Prompt-to-Prompt~\cite{hertz2022prompt} with Null-Text Inversion~\cite{mokady2022null}.}
  \label{fig:5_6}
\end{figure*}
\noindent {\bf Quantitative Results.}
We follow Rerender-A-Video~\cite{yang2023rerender}, VMC~\cite{jeong2023vmc} and MotionDirector~\cite{zhao2023motiondirector} to conduct a user study for quantitative comparison of our method with Ebsynth~\cite{jamrivska2019stylizing}, PikaLabs~\cite{pika} and AnyV2V~\cite{ku2024anyv2v}. We collect videos from \url{https://www.pexels.com/}, DAVIS 2017 dataset~\cite{pont20172017}, UBC Fashion dataset~\cite{zablotskaia2019dwnet} and test videos offered by Stable Video Diffusion~\cite{blattmann2023stable}. These videos cover several categories, including animals, vehicles, and humans. We edit first frames of these videos using EditAnything~\cite{gao2023editanything}, AnyDoor~\cite{chen2023anydoor}, InstantStyle~\cite{wang2024instantstyle} and Instruct-Pix2Pix~\cite{brooks2022instructpix2pix}. These keyframes are used as conditional keyframes for Ebsynth, AnyV2V, and our method. For PikaLabs, we utilize text prompts and extra bounding boxes drawn via their online tools to generate local editing results. 
%Finally, we obtain 10 results of each method for local editing tasks, and 10 results of each method for other tasks, including global style transfer, identity manipulation, and subject customization.
Finally, we obtain 20 results for each method and the editing types include local editing, global style transfer, identity manipulation, and subject customization.
We randomly shuffle the results and display videos to $32$ participants. We ask them to choose the best videos for local editing tasks in four aspects: motion preservation (MP), appearance alignment with source video in the unedited area (AA), overall editing quality (EQ), and temporal consistency (TC). As for other tasks, we ask participants to choose the best videos in the aspect of appearance consistency with the first frame (AC), instead of AA and EQ. We also follow the LOVEU-TGVE competition~\cite{wu2023cvpr} to conduct automatic evaluations, assessing temporal consistency by computing the average CLIP score~\cite{hessel2021clipscore} between frames. The results are shown in \cref{tab:5_1}. Our method achieves the best performance in all aspects of human evaluations and also achieves the best temporal consistency for other tasks of automatic evaluations, demonstrating the superior editing capability of our proposed method. Comparison with the ablation version without fine-grained attention matching (AM) also shows the effectiveness of AM.

\subsection{Ablation Study}
\label{sec:5-3}

\begin{table}[tb]
  \small
\caption{Anderson Normality Test~\cite{anderson1954test} on SVD and SD to evaluate the effectiveness of SARP.}
\vspace{-0.3cm}
\label{tab:5_2}
\centering
\begin{tabular}{@{}cc|cc@{}}
\toprule
\multirow{2}{*}{Model} & \multirow{2}{*}{SARP} & \multirow{2}{*}{\begin{tabular}[c]{@{}c@{}}Statistics$\downarrow$\\ w/ text\end{tabular}} & \multirow{2}{*}{\begin{tabular}[c]{@{}c@{}}Statistics$\downarrow$\\w/o text\end{tabular}} \\
                       &                       &                                                                                   &                                                                                 \\ \midrule
SVD                    & w/                    & -                                                                                & \textbf{91.48}                                                                           \\
SVD                    & w/o                   & -                                                                                 & 2785.10                                                                         \\
SD                     & w/                    & \textbf{1379.82}                                                             & \textbf{944.47}                                                                         \\
SD                     & w/o                   & 3297.29                                                         & 3201.22                                                                         \\ \bottomrule
\end{tabular}
\vspace{-0.2cm}
\end{table}%\vspace{1em}

\noindent {\bf Analysis on Smooth Area Random Perturbation.}
We conduct experiments to evaluate the effectiveness of smooth area random perturbation (SARP). We use test videos from UBC Fashion~\cite{zablotskaia2019dwnet}, perform EDM~\cite{Karras2022edm} inversion on Stable Video Diffusion (SVD)~\cite{blattmann2023stable} with and without SARP. The visual results are generated by EDM denoising using the inverted latents and image prompts, as shown in \cref{fig:5_5}. SARP remarkably improves the inversion results on SVD. SARP is also effective for text-guided image generation model, \eg, Stable Diffusion (SD)~\cite{rombach2021highresolution}, as shown in \cref{fig:5_6}. We conduct experiments on SD with similar settings as SVD, extract keyframes from test videos, perform DDIM~\cite{song2020denoising} inversion, and denoise using original prompts. We also experiment on Prompt-to-Prompt~\cite{hertz2022prompt} with Null-Text Inversion~\cite{mokady2022null}. Results without SARP tend to generate artifacts and fail to produce reasonable edits.
To better assess the quality of inversion, we conduct experiments on a test set of UBC Fashion~\cite{zablotskaia2019dwnet}, which contains $100$ videos with constant white backgrounds. We modify the background color to random constant values and perform Anderson Normality Test~\cite{anderson1954test} on inverted latents to assess their conformity to Gaussian distribution. The quantitative results are shown in \cref{tab:5_2}. Two sets are calculated for SD: one with text prompts generated by BLIP~\cite{li2022blip}, the other without text prompts. Quantitative results demonstrate the effectiveness and generalizability of SARP. 
We include more ablation studies related to SARP in \cref{app:sarp}.

\noindent {\bf Analysis on Skip-Interval Cross-Attention.}
We conduct experiments to compare results with and without skip-interval cross-attention (SIC). The source video is divided into $9$ clips, with motion LoRAs trained for each clip, both with and without SIC. The results are shown in the 3th and 5th row in \cref{fig:5_3}. Results without SIC lose appearance details, especially after several clips as the woman turns back. SIC helps preserve appearance when the keyframe is of low quality. 

\noindent {\bf Analysis on MotionLoRA.} We include ablation study and discussions with Motion LoRA in \cref{app:motionlora}.

\noindent {\bf Analysis on Fine-Grained Attention Matching.}
We compare results with and without fine-grained attention matching (AM), as shown in \cref{fig:5_3}, \cref{fig:5_4} and \cref{tab:5_1}. AM improves motion accuracy and appearance consistency, improving the overall editing quality. We also conduct experiments to demonstrate the impact of different stage partitions of the temporal attention selector on the results, which are included in \cref{app:am_2}.

\section{Conclusion}
\label{sec:6}
In this paper, we propose a novel framework for video editing using a pre-trained image-to-video model. Given an arbitrarily edited first frame, our framework generates edited results for the source video, preserving appearances and motion based on the editing extent. Initially, we align the coarse motion of the output with the source video by training motion LoRAs and employing skip-interval cross-attention to mitigate the quality decline in long video generation. We then refine the appearances and motion of the edited video through fine-grained attention matching, supplemented by smooth area random perturbation for improved editing quality in videos with constant pixel area. Extensive experiments exhibit the effectiveness of our proposed method, which takes a solid step toward extending image editing methods to the video domain.
We discuss the limitations of our method in \cref{app:limitation}.

\small
\bibliographystyle{ieeenat_fullname}
\bibliography{main}

\clearpage
\onecolumn
\section*{Appendix}
%\begin{appendices}
\appendix

\noindent\textbf{Overview.} The appendix includes sections as follows:

\begin{quote}
\begin{itemize}
 \item Ablation and Comparative Study of Smooth Area Random Perturbation (SARP) (Section~\ref{app:sarp}).

    \item Analysis on the Attention matching (Section~\ref{app:am}).

    \item Discussions with Motion Lora (Section~\ref{app:motionlora}).

    \item Additional Visual Results and Comparisons  (Section~\ref{app:add_results}).

    \item Discussions with Limitations  (Section~\ref{app:limitation}).
\end{itemize}
\end{quote}

\section{Smooth Area Random Perturbation}
\label{app:sarp}
% add 1) more results with constant background of other colors, 2) ablation study on non-smooth area 3) add noise in the latent domain and discussion 4) results with different noise scale 
\subsection{Non-Smooth Area and Latent Random Perturbation}
\label{app:sarp_1}
\begin{figure}[htbp]
    \centering
    \includegraphics[width=1.0\textwidth]{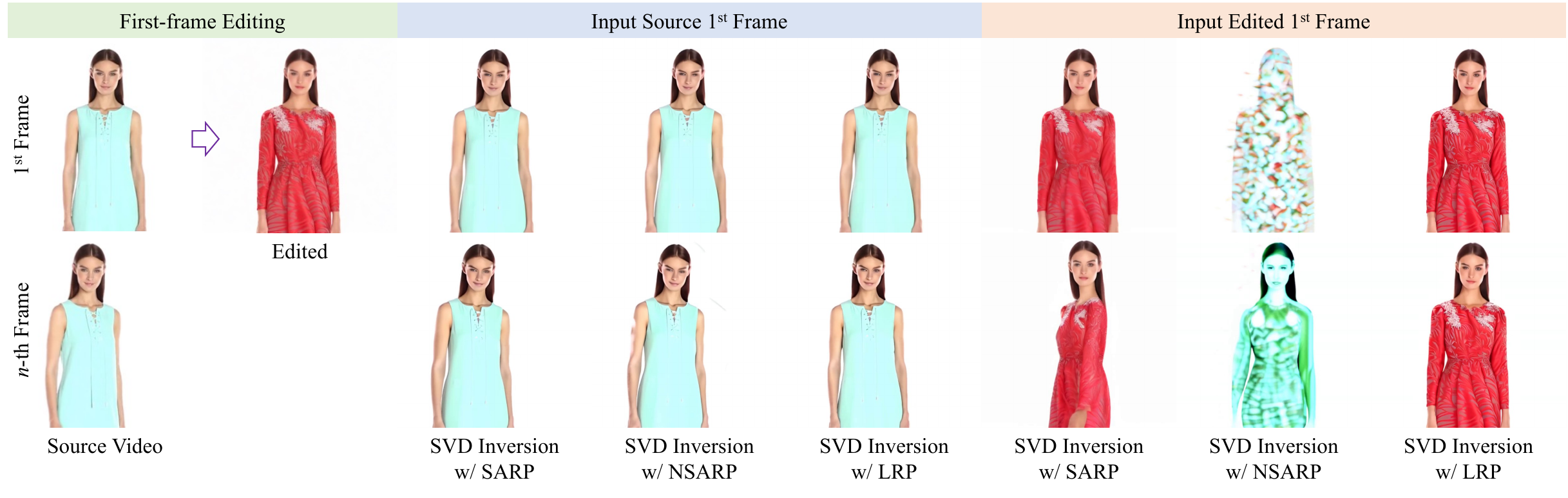}
    \vspace{-0.5cm}
    \caption{Qualitative comparison of SARP, NSARP, and LRP. SVD inversion is integrated with SARP, NSARP, and LRP respectively. While results generated solely by SVD inversion may exhibit motion mismatches with source videos, the integration with SARP or LRP significantly improves the performance compared to those obtained with NSARP.}
    \vspace{-0.1cm}
    \label{fig:app_1_1}
\end{figure}

\begin{figure}[htbp]
    \centering
    \includegraphics[width=1.0\textwidth]{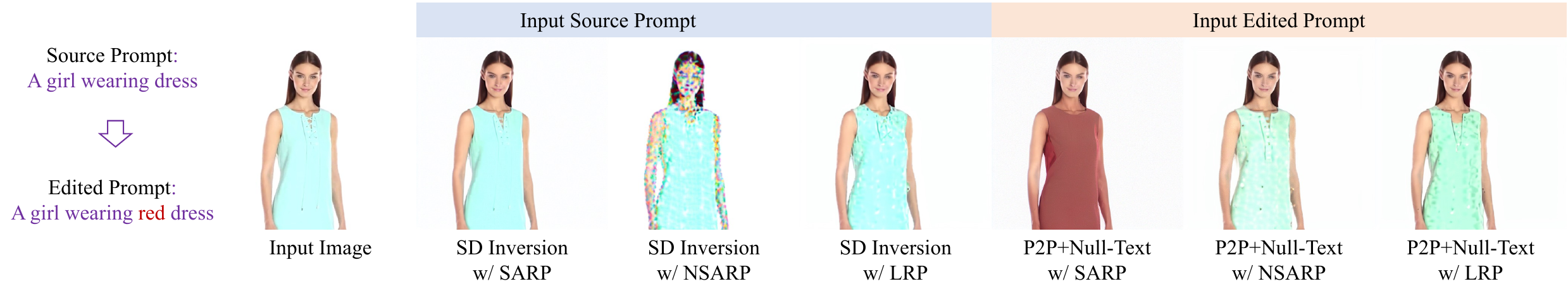}
    \vspace{-0.5cm}
    \caption{Qualitative comparison with NSARP and LRP.}
    \vspace{-0.1cm}
    \label{fig:app_1_2}
\end{figure}

To further evaluate the effectiveness of smooth area random perturbation, we conduct experiments to compare it with the results obtained from non-smooth area random perturbation, \eg, adding noise to the non-smooth area rather than the smooth area. Additionally, we compare the results with those obtained by adding global noise in the latent domain. The results are shown in \cref{fig:app_1_1,fig:app_1_2}. ``NSARP" denotes non-smooth area random perturbation, while ``LRP" refers to latent random perturbation. NSARP produces results with noticeable artifacts and fails to achieve satisfactory editing with Prompt-to-Prompt~\cite{hertz2022prompt}. These results suggest that the artifacts produced without SARP are primarily rooted in the smooth area with constant pixels. LRP diminishes artifacts for SVD~\cite{blattmann2023stable} and SD~\cite{rombach2021highresolution} inversion, but we observe that its performance still falls behind that of SARP, particularly for image editing using Prompt-to-Prompt~\cite{hertz2022prompt} with Null-Text Inversion~\cite{mokady2022null}. We set noise scale $\alpha = 0.005$ for SVD, and $\alpha = 0.02$ for SD.  

\subsection{Constant pixel area with other colors}
\label{app:sarp_2}
\begin{figure}[htbp]
    \centering
    \includegraphics[width=1.0\textwidth]{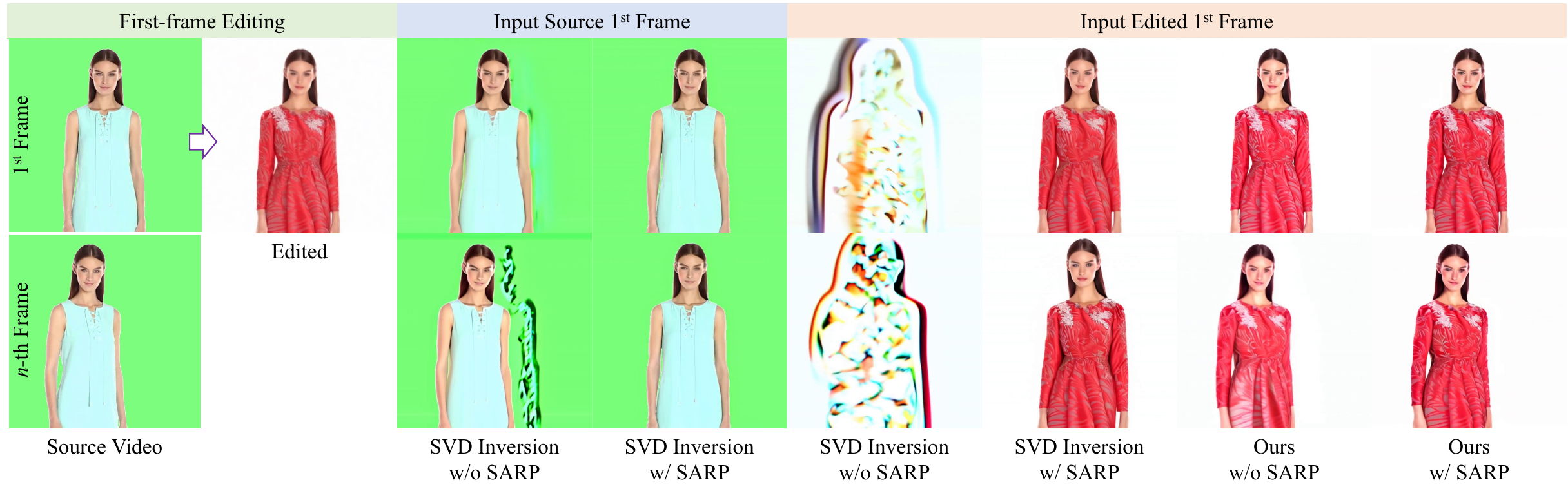}
    \vspace{-0.5cm}
    \caption{Results of inputs with green background.}
    \vspace{-0.1cm}
    \label{fig:app_1_3}
\end{figure}

We offer another group of results generated with the constant area of other colors, as shown in \cref{fig:app_1_3}. The experimental results are consistent with the results of experiments conducted with white background inputs. SARP remarkably improves the results of inversion and editing. The results suggest that artifacts produced without SARP are primarily rooted in the smooth area, and are not related to the colors of these regions.

\subsection{Ablation Study on Noise Scale}
\label{app:sarp_3}
\begin{figure}[htbp]
    \centering
    \includegraphics[width=1.0\textwidth]{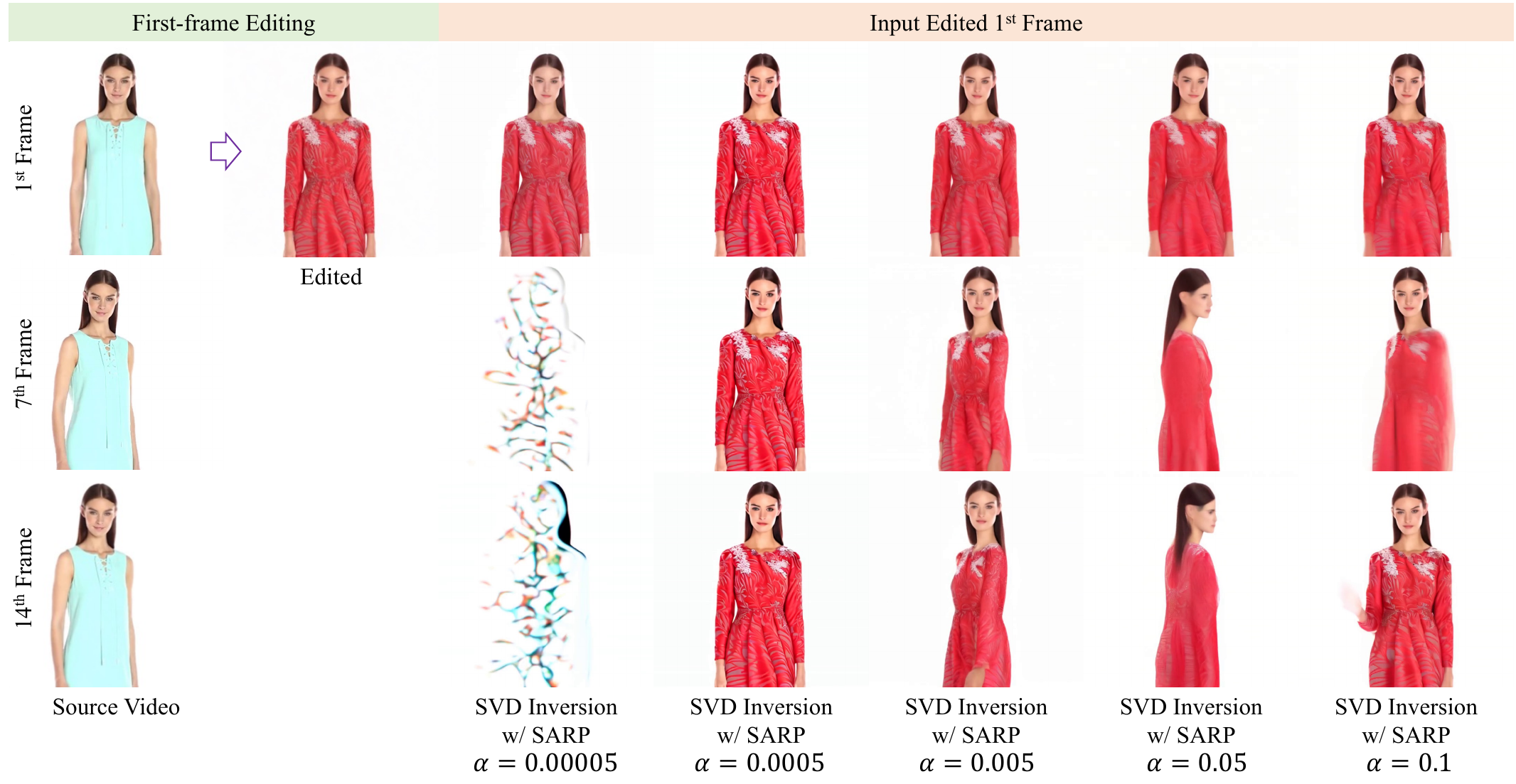}
    \vspace{-0.5cm}
    \caption{Ablation study on noise scale.}
    \vspace{-0.1cm}
    \label{fig:app_1_4}
\end{figure}

We conduct experiments to study the influence of different noise scales of SARP on SVD inversion, as shown in \cref{fig:app_1_4}. Small noise scale, \eg, $\alpha=0.00005$, would reduce the effectiveness of SARP, and generate results with artifacts. A large noise scale, \eg, $\alpha=0.1$, tends to generate motions with more artifacts. We find $\alpha \in [0.0005, 0.005]$  is suitable for producing satisfactory results. We leave further study on SARP for future work.
\label{sec:1_3}

\subsection{Comparison with Shifted Noise}
\label{app:sarp_4}
\begin{figure}[htbp]
    \centering
    \includegraphics[width=1.0\textwidth]{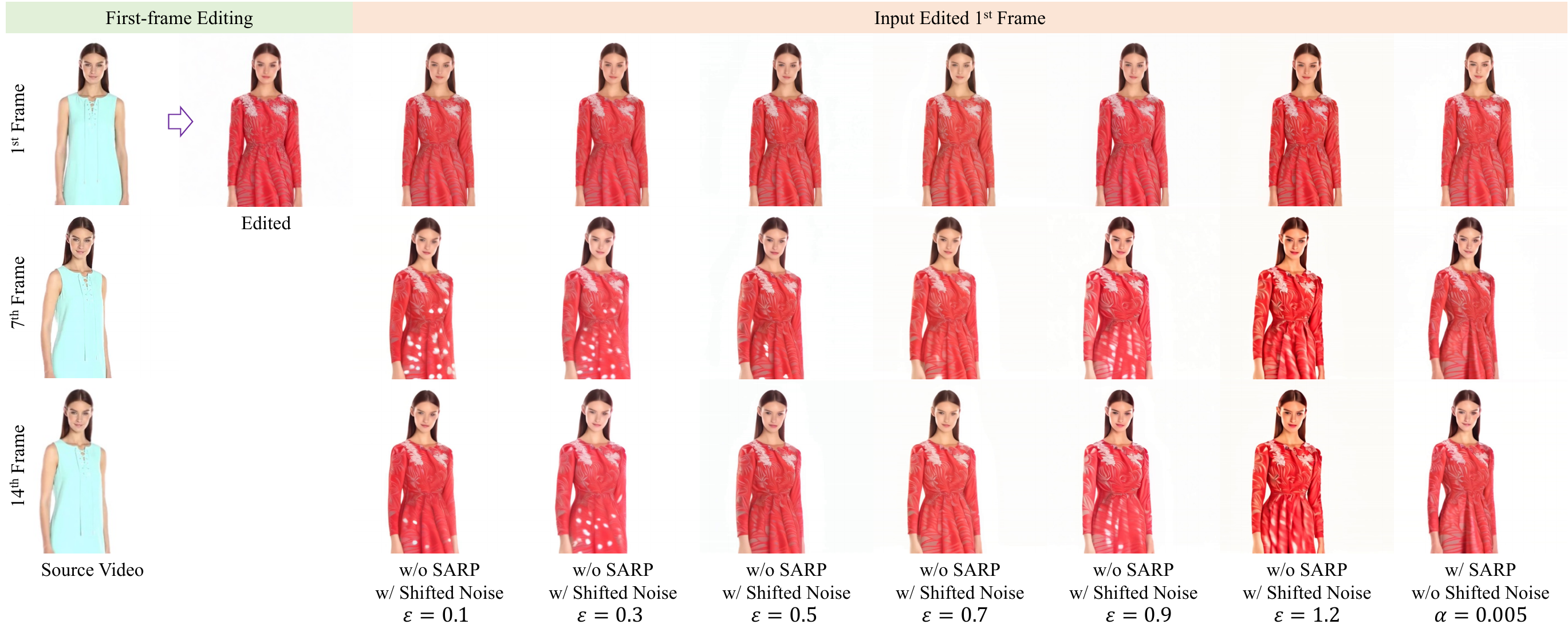}
    \vspace{-0.5cm}
    \caption{Comparison with Shifted Noise.}
    \vspace{-0.1cm}
    \label{fig:app_1_5}
\end{figure}

\begin{figure}[htbp]
    \centering
    \includegraphics[width=0.8\textwidth]{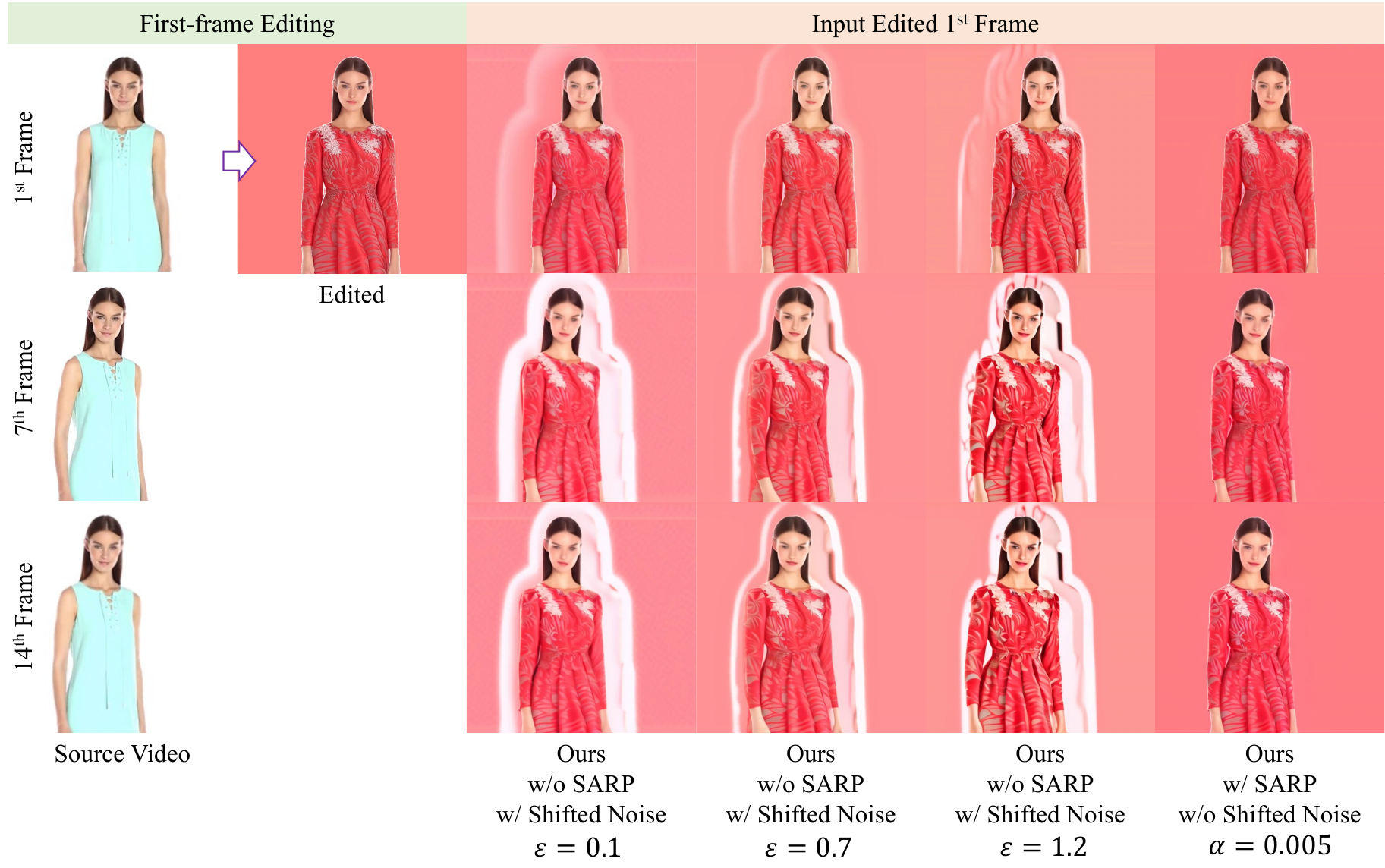}
    %\vspace{-0.5cm}
    \caption{Comparison with Shifted Noise.}
    %\vspace{-0.1cm}
    \label{fig:app_1_6}
\end{figure}

We conduct experiments to compare SARP with shifted noise~\cite{shiftednoise} for handling images with large smooth areas. We generate outputs using models trained with different shifted noise scales (without SARP), from $\epsilon=0.1$ to $\epsilon=1.2$. As shown in \cref{fig:app_1_5}, results of shifted noise exhibit severe artifacts, while SARP demonstrates better performance. Additionally, SARP generalizes better to images with different background colors compared to source videos, as shown in \cref{fig:app_1_6}.

\section{Fine-Grained Attention Matching}
\label{app:am}
% add 1) visualization of difference map to show clearly the effectiveness of difference map
\subsection{Spatial Self-Attention Maps}
\label{app:am_1}
\begin{figure}[htbp]
    \centering
    \includegraphics[width=1.0\textwidth]{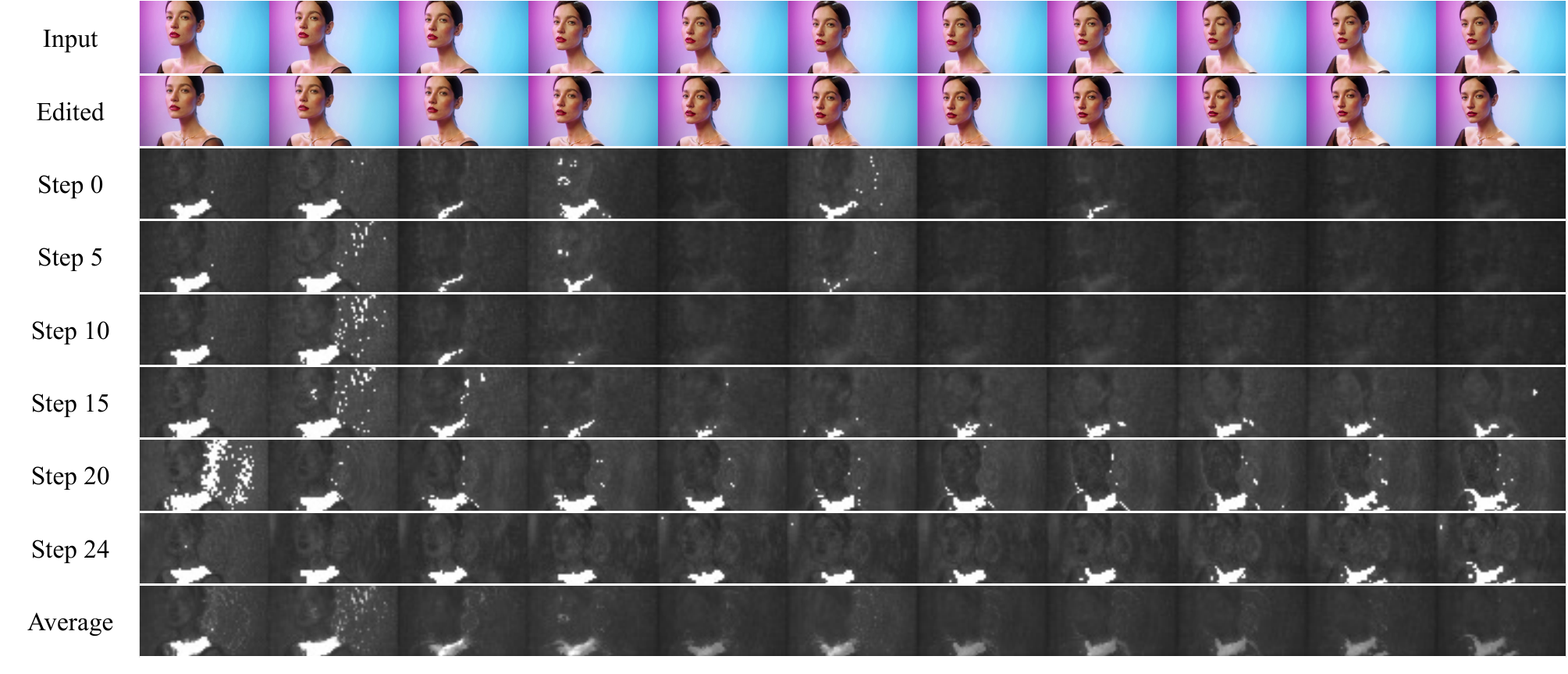}
    \vspace{-0.5cm}
    \caption{Visualisation of spatial self-attention difference maps, generated by the third layer of downscale modules.}
    \vspace{-0.1cm}
    \label{fig:app_2_1}
\end{figure}

We visualize spatial self-attention difference maps of the third layer of downscale modules for steps $[0,5,10,15,20,24]$, along with the average maps for all the steps, as shown in \cref{fig:app_2_1}.  White areas indicate the different structure, \eg, the necklace, between original frames and edited frames. This demonstrates that the difference maps can serve as an effective tool to localize the edited region and control the strength of attention matching accordingly.

\subsection{Temporal Attention Selector}
\label{app:am_2}
\begin{figure}[htbp]
    \centering
    \includegraphics[width=1.0\textwidth]{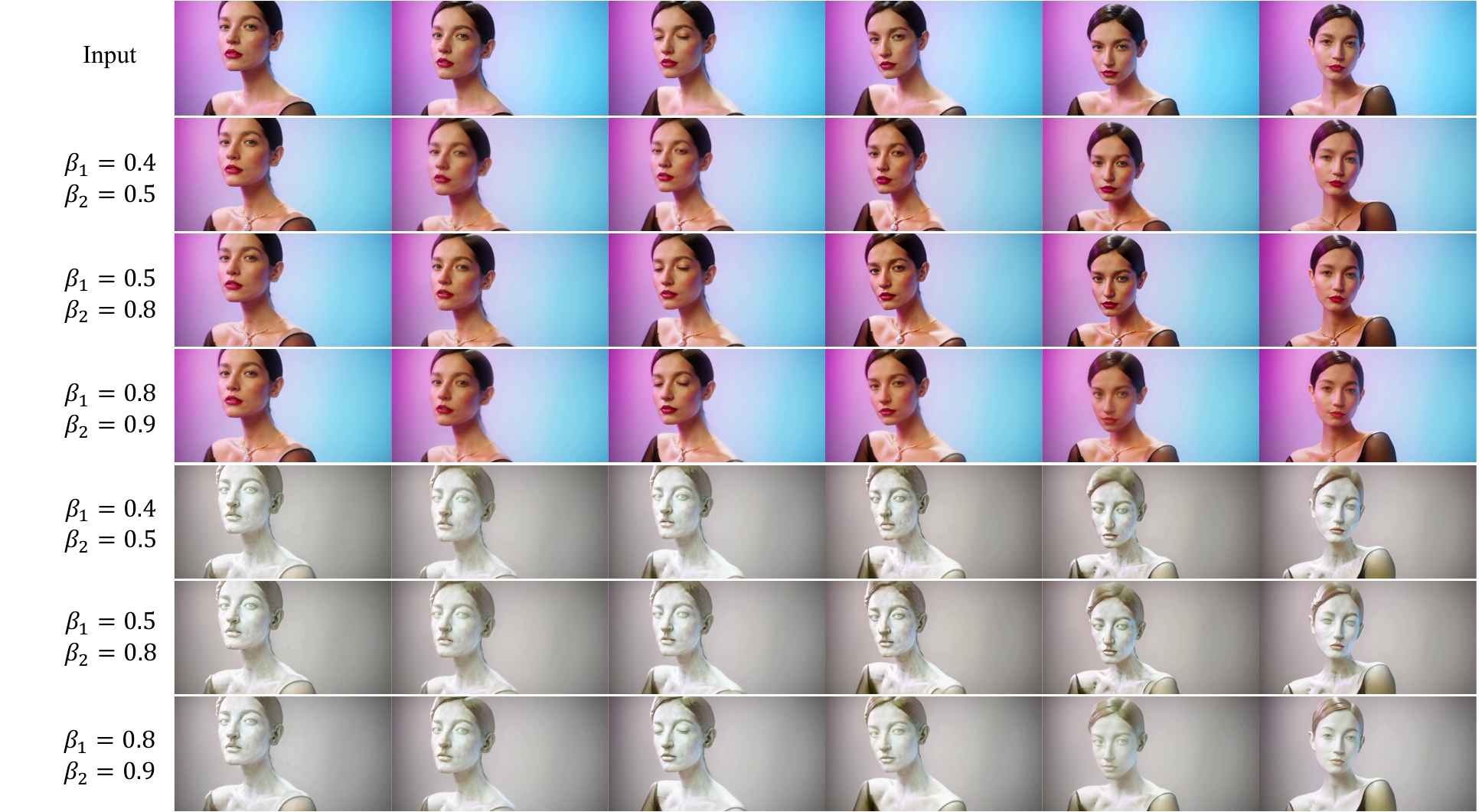}
    \vspace{-0.5cm}
    \caption{Results generated with different stage partitions of temporal attention selector.}
    \vspace{-0.1cm}
    \label{fig:app_2_2}
\end{figure}

\begin{figure}[htbp]
    \centering
    \includegraphics[width=1.0\textwidth]{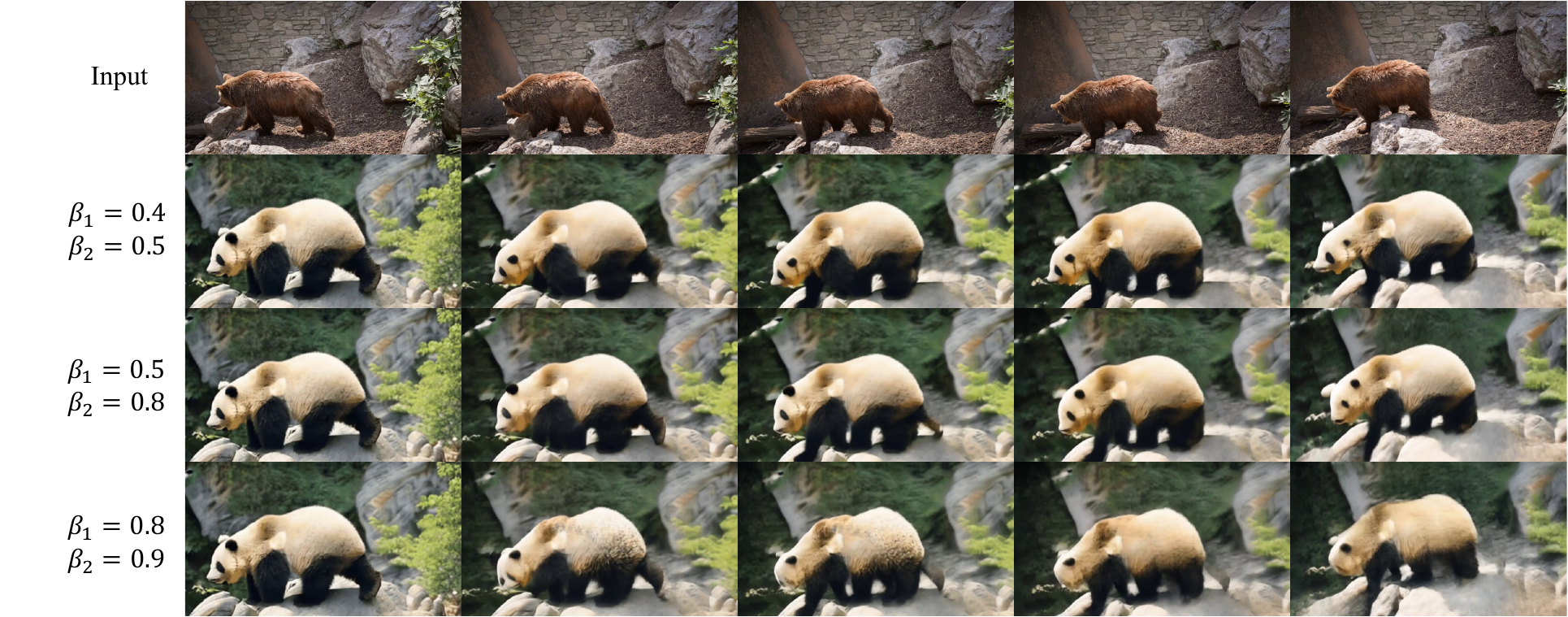}
    \vspace{-0.5cm}
    \caption{Results generated with different stage partitions of temporal attention selector.}
    \vspace{-0.1cm}
    \label{fig:app_2_3}
\end{figure}

We also conduct experiments to analyze the impact of stage partitions of the temporal attention selector on the generated results, as shown in \cref{fig:app_2_2,fig:app_2_3}. For local editing tasks, such as adding a necklace to a woman, the optimal stage partition is $\beta_{1}=0.5,\beta_{2}=0.8$. This setting preserves the structure of the necklace while maintaining the original motions. For style transfer, such as transforming the woman into a Greek sculpture style, the best setting is $\beta_{1}=0.8,\beta_{2}=0.9$. Other settings may result in mismatched motions. For edits involving dramatic structural changes, such as turning a bear into a giant panda, the optimal setting is $\beta_{1}=0.4,\beta_{2}=0.5$. Other settings may cause gradual structural leakage. Since temporal attention captures the optical flow and structural motions of the source video, the best stage partitions depend on the extent of the editing, such as the similarity between the source and edited videos. In our experiments, we classify video editing into three categories: local object editing, edits with dramatic shape changes, and global style transfers. We fix the stage partitions for these three cases, as detailed in \cref{sec:5-1}.

\section{Discussions with Motion LoRA}
\label{app:motionlora}
%3) comparison with other training settings, like training with spatial LoRAs, etc.
\subsection{Discussions with Training}
\begin{figure}[htbp]
    \centering
    \includegraphics[width=0.8\textwidth]{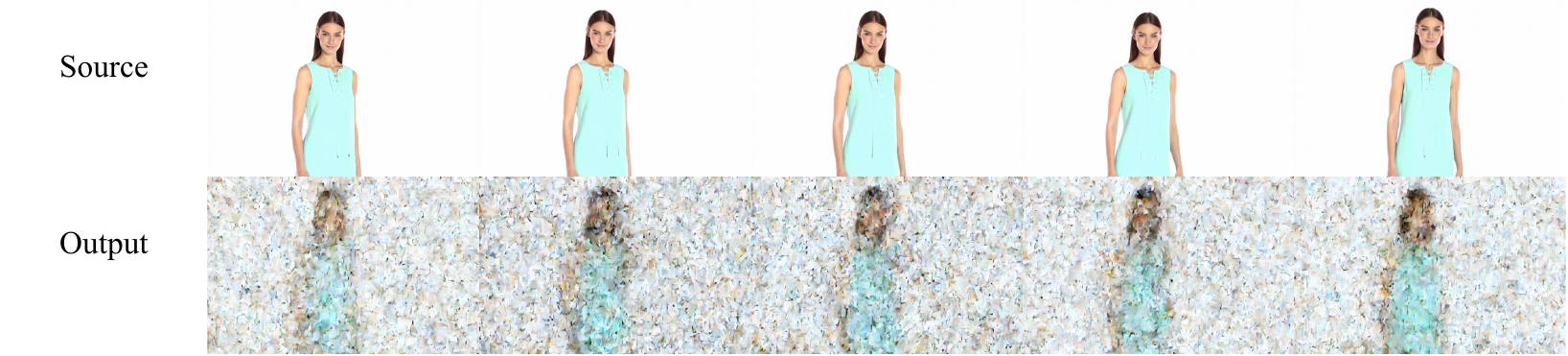}
    %\vspace{-0.5cm}
    \caption{Results generated by model trained with the appearance-debiased loss.}
    \vspace{-0.1cm}
    \label{fig:app_3_1}
\end{figure}

We conduct experiments to train Motion LoRAs on SVD using the same training settings as MotionDirector~\cite{zhao2023motiondirector}.  For spatial LoRA training, we implement two versions: training with randomly selected frames from source video, and training with the edited image. We find that after training for several iterations, the model tends to generate invalid outputs, \eg, frames with pixel values of NaN. We also find that the model trained with the appearance-debiased loss proposed by MotionDirector generates unsatisfactory outputs, as shown in \cref{fig:app_3_1}. We abandon the spatial LoRAs and the appearance-debiased loss in our main framework.

\subsection{Ablation Study on Motion LoRA}
\begin{figure}[htbp]
    \centering
    \includegraphics[width=1.0\textwidth]{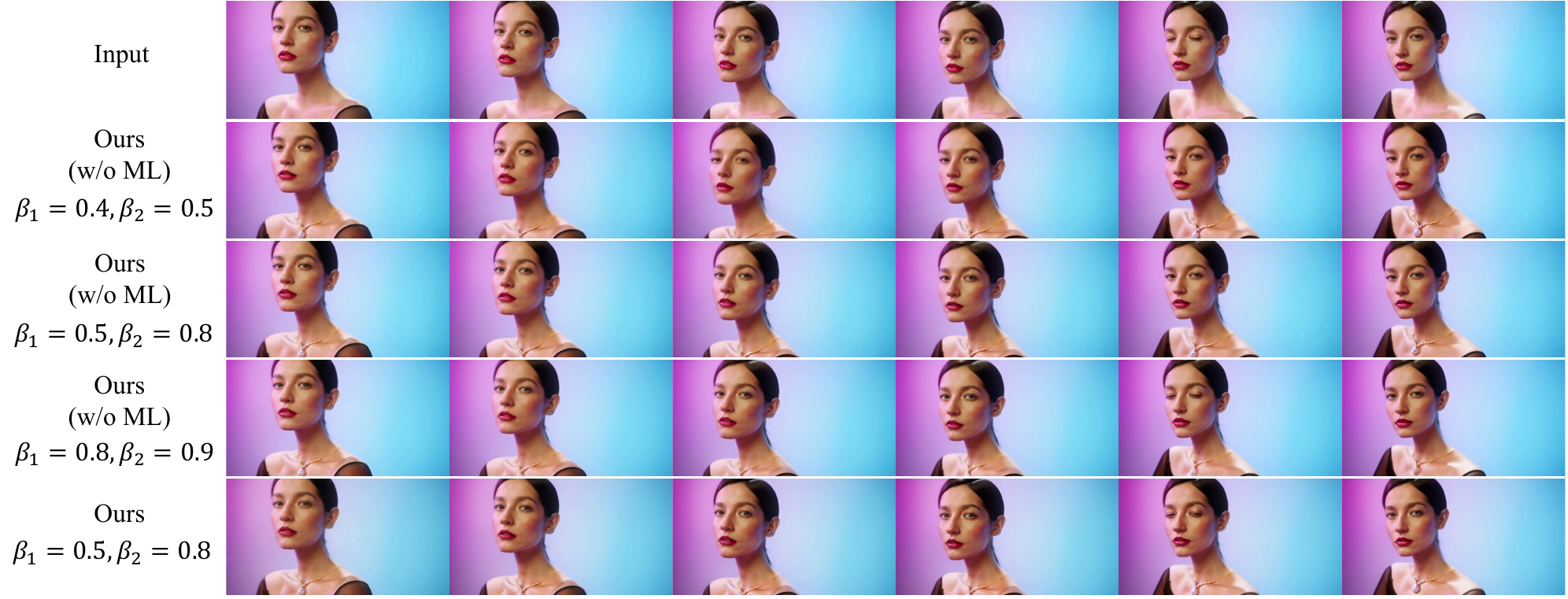}
    %\vspace{-0.5cm}
    \caption{Ablation study on Motion LoRA (ML).}
    \vspace{-0.1cm}
    \label{fig:app_3_2}
\end{figure}

\begin{figure}[htbp]
    \centering
    \includegraphics[width=1.0\textwidth]{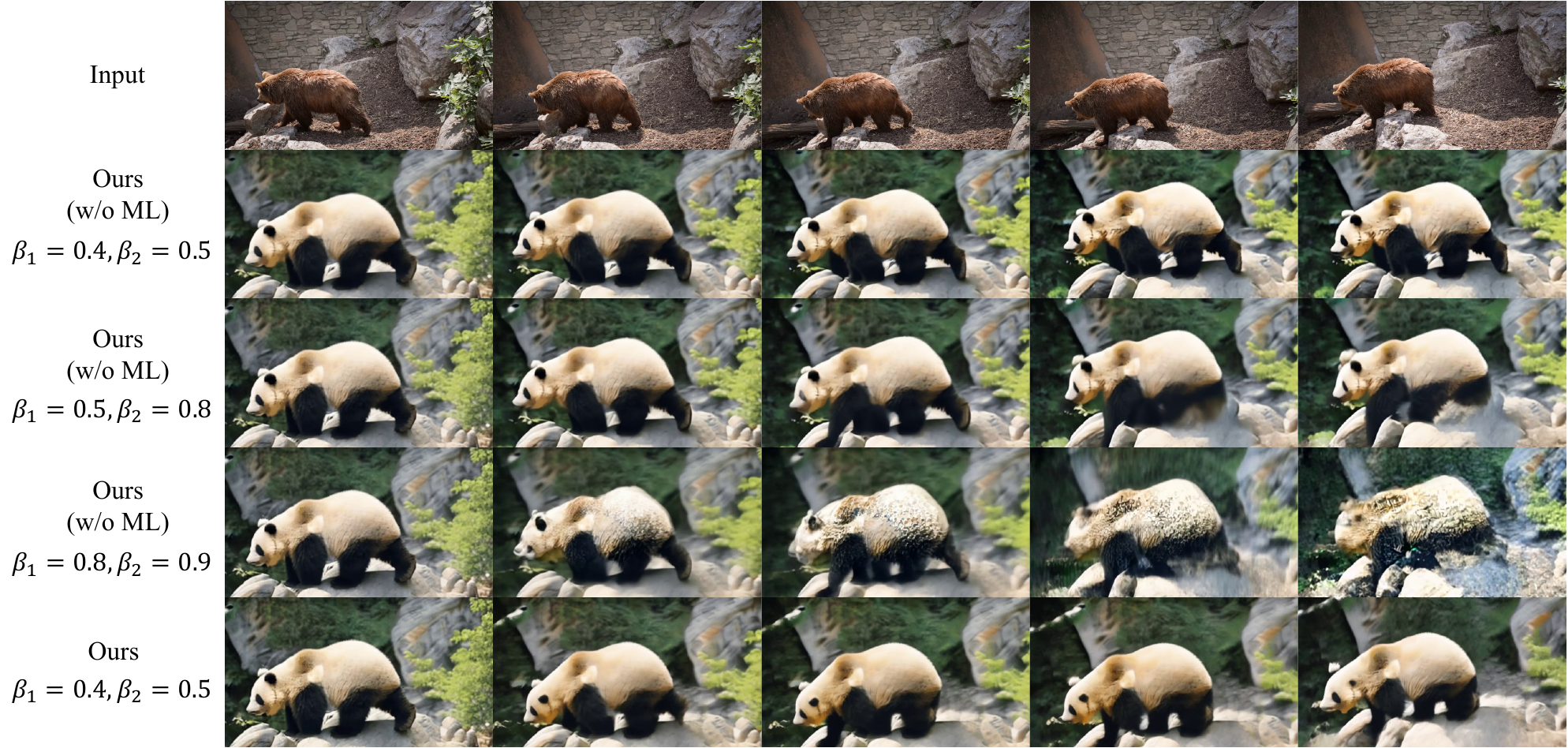}
    %\vspace{-0.5cm}
    \caption{Ablation study on Motion LoRA (ML).}
    \vspace{-0.1cm}
    \label{fig:app_3_3}
\end{figure}

\begin{figure}[htbp]
    \centering
    \includegraphics[width=1.0\textwidth]{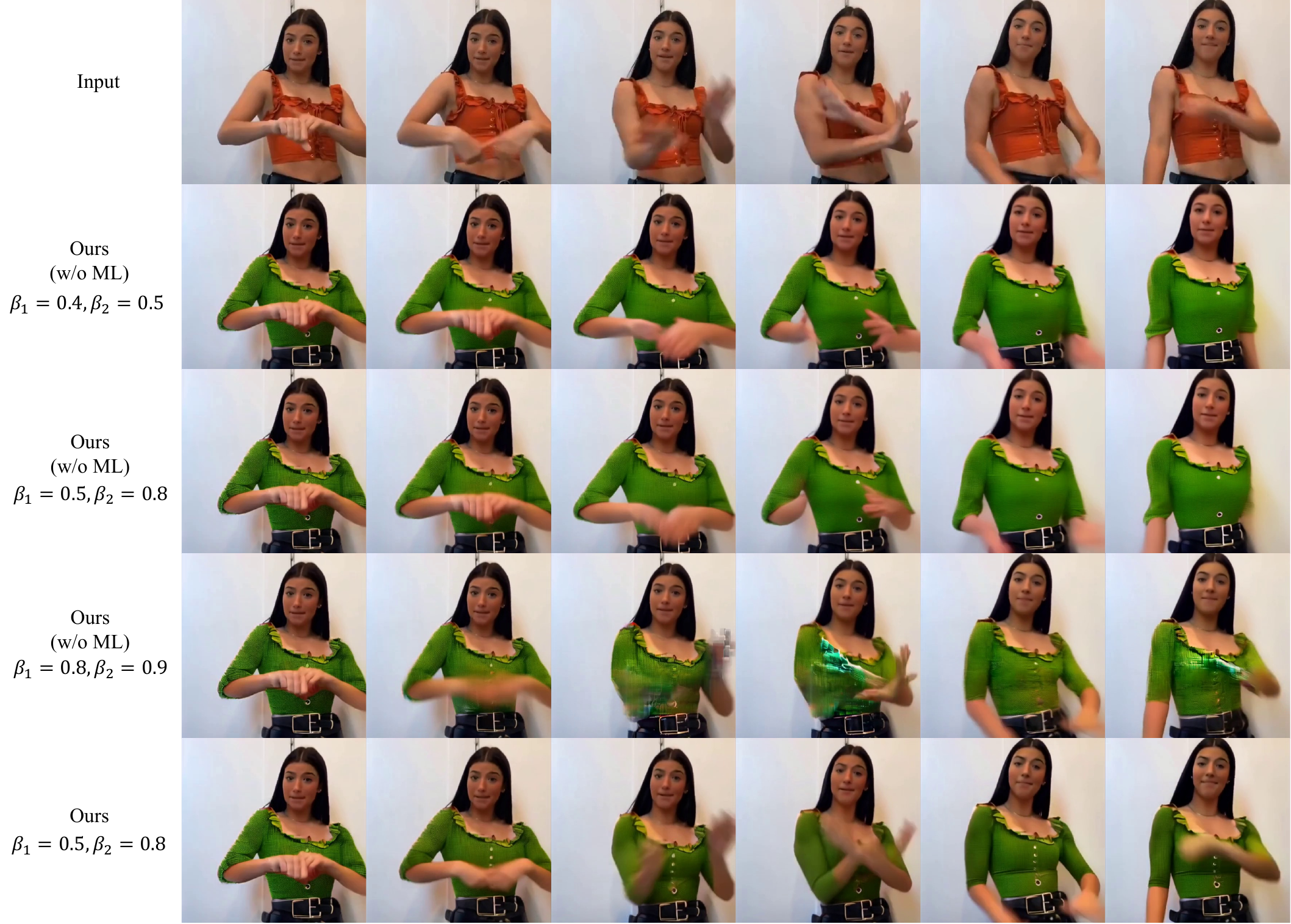}
    %\vspace{-0.5cm}
    \caption{Ablation study on Motion LoRA (ML).}
    \vspace{-0.1cm}
    \label{fig:app_3_4}
\end{figure}

We conduct experiments to evaluate the effectiveness of Motion LoRA (ML), as shown in \cref{fig:app_3_2,fig:app_3_3,fig:app_3_4}. Results generated without Motion LoRA fail to match the motions of the source video, \eg, the girl fails to wink her eyes in the 5th column of \cref{fig:app_3_2}. This issue is more pronounced when editing involves significant structural changes, as seen in \cref{fig:app_3_3}, and when editing videos with rapid motions, as shown in \cref{fig:app_3_4}. In these cases, the absence of Motion LoRA tends to magnify the motion mismatch between source and edited videos, as well as introduce artifacts, regardless of how the stage partitions of the temporal attention selector are chosen in attention matching.

\section{Other Comparisons}
\label{app:add_results}
\begin{figure}[htbp]
  \centering
  \includegraphics[width=1.0\textwidth]{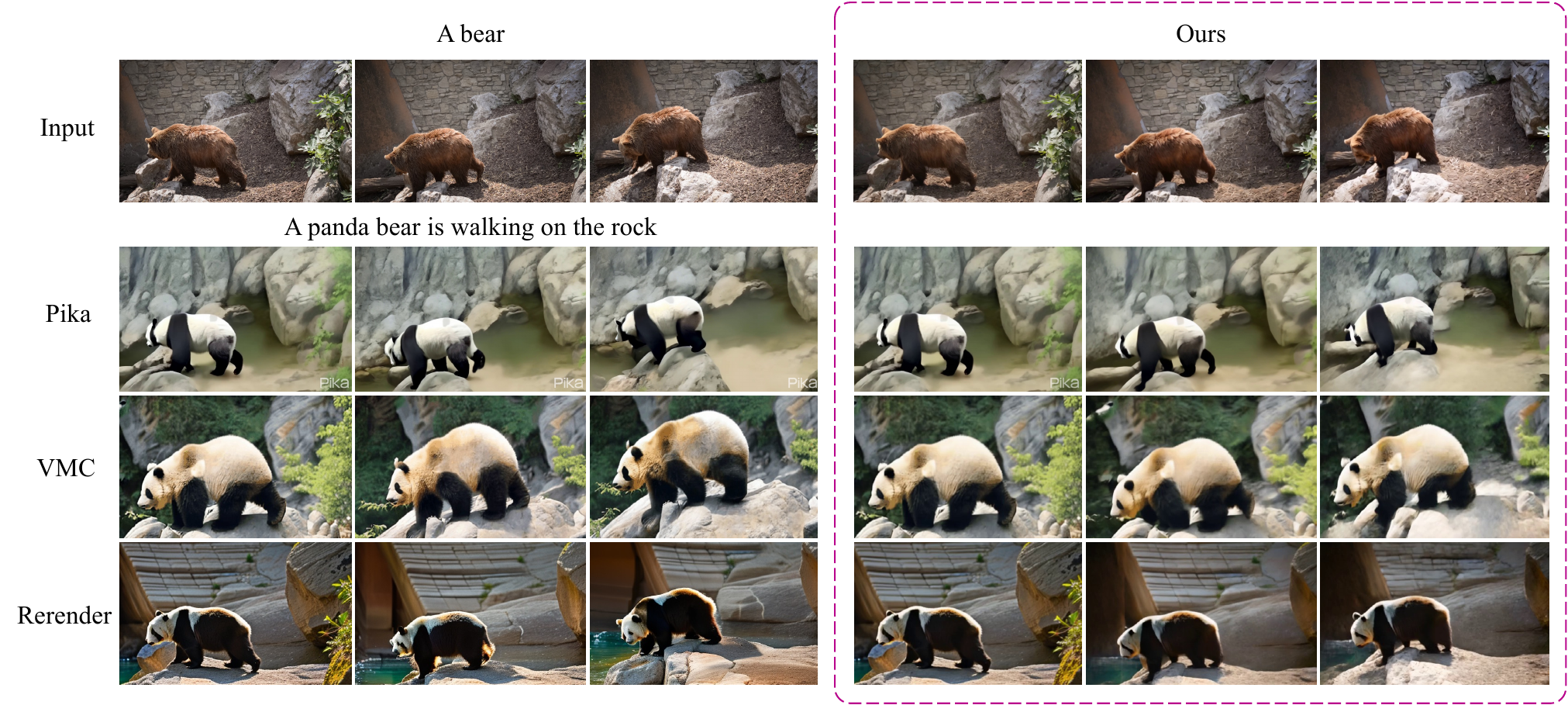}
  \vspace{-0.5cm}
  \caption{Qualitative comparison with text-guided video editing and motion customization. We use the first frames generated by these methods as conditional keyframes for our method to generate editing results.}
  \vspace{-0.1cm}
  \label{fig:app_5_2}
\end{figure}

We offer another visual comparison with Rerender-A-Video~\cite{yang2023rerender}, VMC~\cite{jeong2023vmc} and PikaLabs~\cite{pika}, as shown in \cref{fig:app_5_2}. We use the first frames generated by these text-guided methods as initial keyframes to generate editing results, which shows the capability of our method to handle global editing and style transfer tasks. For more results. please visit our website at \url{https://i2vedit.github.io/}.

\section{Limitations}
\label{app:limitation}
\iffalse
Although our framework can adaptively preserve appearances and motion from source video based on the editing extent without any extra masks, there are still some cases where the model generates results with color and texture slightly different from source video in unedited areas. %It will more perfectly preserve the appearances to utilize an extra mask for our model, like PikaLabs~\cite{pika} does. 
This could be addressed by using an extra mask for our model, as PikaLabs~\cite{pika} does. 
%Furthermore, although skip-interval cross attention could reduce quality decline to some extent, we find that the quality of edited results will suffer from severe decline after three or four iterations if the source video contains more motion and appearance informations than the base image-to-video model could generate. 
Furthermore, although skip-interval cross-attention helps to preserve the frame quality, we find that the edited results may gradually deviate from the edited first frame after about four clip iterations, as the effects of skip-interval cross-attention get weaker for increasingly different video content after a long time range. %if the source video contains more motion and appearance information than the base image-to-video model could generate.
%In these cases, new content generated after a long time range is very different from the first clip, skip-interval cross attention doesn't play a significant role.
We leave longer video editing as a future work.
\fi
Although our framework can adaptively preserve appearances and motion from source video based on the editing extent without any extra masks, there are still some cases where the model generates results with color and texture slightly different from source video in unedited areas. %It will more perfectly preserve the appearances to utilize an extra mask for our model, like PikaLabs~\cite{pika} does. 
This could be addressed by using an extra mask for our model, as PikaLabs~\cite{pika} does. 
%Furthermore, although skip-interval cross attention could reduce quality decline to some extent, we find that the quality of edited results will suffer from severe decline after three or four iterations if the source video contains more motion and appearance informations than the base image-to-video model could generate. 
%Furthermore, although skip-interval cross-attention helps to preserve the frame quality, we find that the edited results may gradually deviate from the edited first frame after about four clip iterations, as the effects of skip-interval cross-attention get weaker for increasingly different video content after a long time range.
Additionally, on an NVIDIA A100 GPU, training for coarse motion extraction takes about 25 minutes for 250 iterations on a single clip. The appearance refinement pipeline then takes approximately 10 minutes to generate the final outputs. %The exact time depends on the computing resources.
Furthermore, although skip-interval cross-attention helps to preserve the frame quality, we find that the quality of edited results may degrade when the video has substantial content change, \eg, from the front view of a girl to the back view, as shown in \cref{fig:5_3}. %which is likely to happen for long videos. %deviate from the edited first frame after about four clip iterations, as the effects of skip-interval cross-attention get weaker for increasingly different video content after a long time range.
This is because the effects of skip-interval cross-attention would get weaker for increasingly different video content.
%if the source video contains more motion and appearance information than the base image-to-video model could generate.
%In these cases, new content generated after a long time range is very different from the first clip, skip-interval cross attention doesn't play a significant role.
We leave video editing with stronger content change, along with reducing time costs as a future work.

%\end{appendices}

\end{document}